\RequirePackage{fix-cm}
\documentclass[smallextended]{svjour3}       % onecolumn (second format)
\smartqed  % flush right qed marks, e.g. at end of proof
\usepackage{graphicx}
\usepackage{amsmath,amssymb}
\usepackage{booktabs}
\usepackage{chngpage}
\usepackage{lipsum}
\usepackage{caption}
\usepackage{subfig}
\usepackage{hyperref}
\usepackage{threeparttable}
\usepackage{soul}
\usepackage{lineno}
\usepackage{multirow}
\usepackage{natbib}
\usepackage[table]{xcolor}

\begin{document}
\newcommand{\tabincell}[2]{\begin{tabular}{@{}#1@{}}#2\end{tabular}}
\newcommand{\kun}[1]{\textcolor{black}{#1}}
\setcounter{secnumdepth}{0}

\title{Deep Learning Techniques for In-Crop Weed Recognition in Large-Scale Grain Production Systems: A Review%\thanks{Grants or other notes
}

\titlerunning{Deep Learning for Weed Recognition: A Review}        % if too long for running head

\author{Kun~Hu\textsuperscript{1,*}         \and
        Zhiyong~Wang\textsuperscript{1} \and
        Guy~Coleman\textsuperscript{2} \and
        Asher~Bender\textsuperscript{3} \and
        Tingting~Yao\textsuperscript{4} \and
        Shan~Zeng\textsuperscript{5} \and
        Dezhen~Song\textsuperscript{6} \and
        Arnold~Schumann\textsuperscript{7} \and
        Michael~Walsh\textsuperscript{2} %etc.
}

\authorrunning{Kun Hu et al.} % if too long for running head

\institute{* Corresponding Author \at
              \email{hukun\_sdu@hotmail.com}           %  \\
           \and
           1 School of Computer Science, The University of Sydney, NSW 2006, Australia \\
           2 School of Life and Environmental Sciences, The University of Sydney, \\NSW 2006, Australia \\
           3 Australian Centre for Field Robotics, The University of Sydney, NSW 2008,Australia \\
           4 College of Information Science and Technology, Dalian Maritime University, \\Dalian 116000, China \\
           5 College of Mathematics and Computer Science, Wuhan Polytechnic University, \\Hubei 430023, China \\
           6 Department of Computer Science and Engineering, Texas A\&M University, \\College Station, TX 77843-3127, USA \\
           7 Citrus Research and Education Center, University of Florida, Florida 33850-2299, USA
}

\date{Received: date / Accepted: date}

\maketitle

\begin{abstract}
Weeds are a significant threat to agricultural productivity and the environment. 
The increasing demand for sustainable weed control practices has driven innovative developments in alternative weed control technologies aimed at reducing the reliance on herbicides. The barrier to adoption of these technologies for selective in-crop use is availability of suitably effective weed recognition. With the great success of deep learning in various vision tasks, many promising image-based weed detection algorithms have been developed. 
This paper reviews recent developments of deep learning techniques in the field of image-based weed detection. The review begins with an introduction to the fundamentals of deep learning related to weed detection. Next, recent advancements in deep weed detection are reviewed with the discussion of the research materials including public weed datasets. 
Finally, the challenges of developing practically deployable weed detection methods are summarized, together with the discussions of the opportunities for future research.
We hope that this review will provide a timely survey of the field and attract more researchers to address this inter-disciplinary research problem.
\end{abstract}

%\linenumbers

\section{Introduction}
\kun{Due to their competition with crops for water, nutrients and sunlight, weeds are a significant threat to agricultural productivity \citep{llewellyn2016impact,gharde2018assessment}. Weed control in conservation cropping systems is reliant on the use of herbicides due to the lack of suitable alternative weed control options that do not interfere with the principles of minimum tillage and residue retention and thus the substantial benefits of this system. The site-specific approach to weed control (SSWC) creates the opportunity to alleviate this threat through the precision application of alternative weed control treatments such as lasers, electrical weeding, waterjet cutting etc. \citep{coleman2019energy}. However, to achieve selective in-crop weed control, that avoids crop damage with these alternative treatments accurate weed recognition in all conditions is essential.}

\kun{The complex and highly variable cropping environment is a significant barrier to the development of robust weed recognition algorithms \citep{olsen2019deepweeds}. Plant morphologies as influenced by genetics and the environment vary considerably both between plant species (crop and weed) as well as within these species to create a substantial challenge to the development of weed recognition algorithms. It is possible that extent of morphological variability and changing complexity will require differing weed recognition approaches based for example on species (crop and weed), growth stage, environment and combinations of these influences on plant growth. In scenarios where there are substantial differences in plant morphology the weed recognition challenge will be simpler than those where the crop and weed plants are very similar. The majority of previous weed recognition methods were based on conventional computer vision and machine learning techniques in both colour and multispectral imagery. These approaches often follow a pipeline in which hand-crafted image features play a primary role (e.g., shape features \citep{charters2014eagle}). As a result, developing such pipelines is labour intensive and images are required to be captured within well-defined conditions.}

%\begin{figure}[!htb]
%\centering
%\includegraphics[width=0.49\textwidth]{comparision_ml_dl.jpg}
%\caption{Illustration of the comparison between the pipelines of the conventional machine learning and deep learning.}
%\label{fig:comparison_ml_dl}
%\end{figure}

\begin{figure*}[htbp]
\begin{minipage}[c]{0.55\textwidth}
\centering
\includegraphics[width=\textwidth]{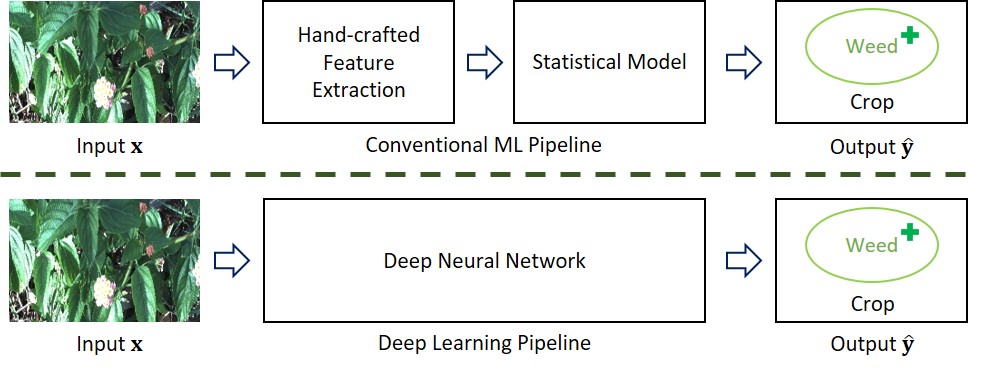}
\end{minipage}
\begin{minipage}[c]{0.43\textwidth}
\caption{Illustration of the comparison between the pipelines of the conventional machine learning and deep learning.}
\end{minipage}
\label{fig:comparisons}
\end{figure*}

\begin{figure*}[htbp]
\begin{minipage}[t]{0.235\textwidth}
\centering
\includegraphics[width=\textwidth]{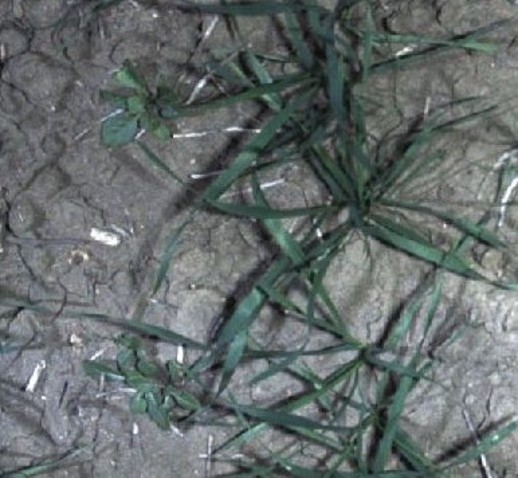}
\caption*{(a) }
\end{minipage}
\begin{minipage}[t]{0.235\textwidth}
\centering
\includegraphics[width=\textwidth]{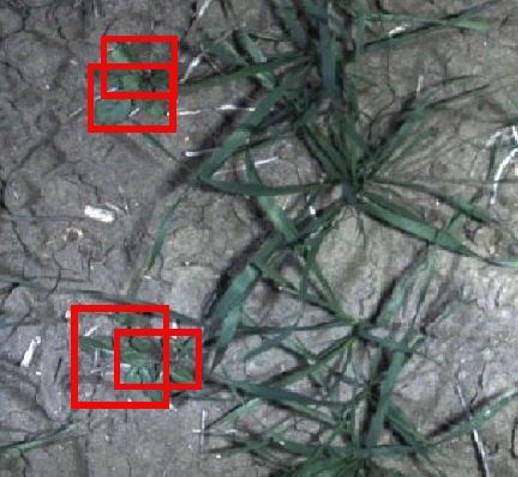}
\caption*{(b)}
\end{minipage}
\begin{minipage}[t]{0.235\textwidth}
\centering
\includegraphics[width=\textwidth]{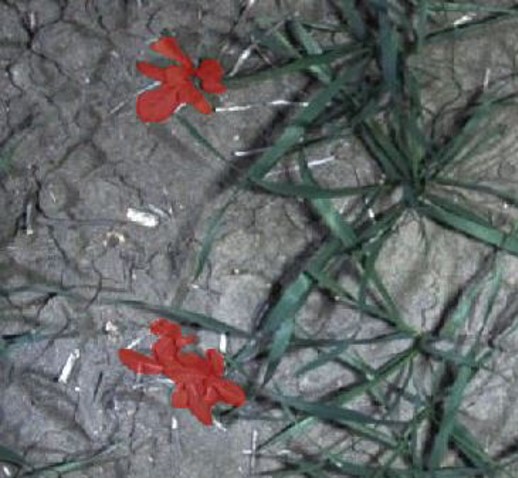}
\caption*{(c)}
\end{minipage}
\begin{minipage}[t]{0.235\textwidth}
\centering
\includegraphics[width=\textwidth]{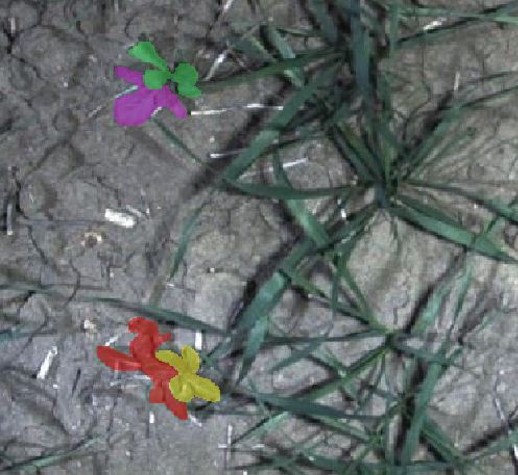}
\caption*{(d)}
\end{minipage}
\caption{Illustration of the four major approaches for weed detection: (a) image classification (b) object detection  (c) semantic segmentation (d) instance segmentation.}
\label{fig:methods}
\end{figure*}

Fortunately, due to the great success of deep learning in many vision tasks, hand-crafted features are no longer required to derive promising results. Instead, deep representations of an input image can be obtained using deep learning, which are relevant to the task at hand. For weed recognition, four types of deep learning approaches are used as illustrated in Fig. \ref{fig:methods}: a) image classification identifies the weed or the crop species contained in an image; b) object detection identifies the per-weed locations of the plants within an image;  c) semantic segmentation conducts pixel-wise classification of individual weed classes and d) instance segmentation further identifies the instance each pixel belongs to. 
As most of the deep learning-based weed recognition studies are based on existing and well-known deep architectures, those relevant to weed recognition, including their building blocks and contributions, are introduced first.
Next, more than 30 deep learning-based weed recognition studies are discussed in terms of their architectures, goals and performance. In addition, as deep learning based weed recognition research often requires a large volume of annotated data, we provide the details of current publicly available weed datasets and benchmarking metrics. 

Expanding on the research of existing well-known architectures, we present other fine-grained and alternative architectures that may offer advantages in terms of the recognition performance for future research. In practice, there are still limitations and challenges for current weed recognition research to provide weed control in large-scale crop production systems. Therefore, deep learning mechanisms which could further improve the efficiency and effectiveness of weed control including real-time inference, weakly-supervised learning, explainable learning and incremental learning techniques are discussed. 

%In conservation cropping systems, herbicides are relied on almost exclusively for weed management. To alleviate risks of herbicide resistance though there are moves to implement an integrated approach to weed management (IWM) that incorporates additional forms of weed control such as crop competition \citep{walsh2014hwsc} and harvest weed seed control \citep{chauhan2012ecology}. Nevertheless, the low cost, high selectivity, ease of application, immediacy of control and lack of viable alternatives ensures the ongoing reliance on herbicides in global cropping systems. 

%In addition, accurate weed recognition and site-specific application of treatments substantially reduces the environmental footprint of herbicides compared to the current whole field, indiscriminate treatment method \citep{lottes2018joint}.

In summary, this review aims to: 1) investigate deep learning techniques related to weed control; 2) summarize current deep learning-based weed recognition research including architectures, research materials and evaluation methods; 3) identify further challenges and improvements for future research with deep learning-based solutions.

The remainder of the review is organized as follows. In Section \textit{Overview of Deep Learning Techniques}, deep learning architectures related to weed control are introduced. Section \textit{Deep Learning for Weed Recognition} provides a discussion of the existing deep learning based weed detection studies. In addition, public datasets and evaluation metrics for benchmarking are summarised. Section \textit{Discussion} considers the challenges for weed detection and potential deep learning solutions. 
Finally, Section~\textit{Conclusion} summarises this review.

\section{Overview of Deep Learning Techniques}
\label{sec:deep}

In this section, the theory of deep learning techniques is introduced %, which have achieved promising performance for a wide range of applications. %Machine learning is introduced before different deep learning techniques are discussed.
including the deep learning building blocks and architectures relevant for weed detection.% are discussed. 

\subsection{Machine Learning}

Machine learning (ML) algorithms are a class of algorithms that 'learn' to perform a specific task given sample data (i.e., training data). These algorithms are not explicitly programmed with rules or instructions to fulfil the task. 
In general, a set of samples for a ML task $\mathbf{D}=\{(\mathbf{x}_{i}, \mathbf{y}_{i})\}$ can be obtained, where $\mathbf{x}_{i} \in \mathbf{R}^{p}$ is the observed features describing the characteristics of the $i$-th sample and $\mathbf{y}_{i}$ is the associated output. In general, for $(\mathbf{x}, \mathbf{y})\in \mathbf{D}$, it can be costly and time-consuming to obtain $\mathbf{y}$, whilst $\mathbf{x}$ is convenient to collect. Therefore, it is expected to learn a model $f_{\mathbf{\Theta}}(\mathbf{x})$ that maps input values to the target variable $\hat{\mathbf{y}}$ as close as possible, where $\mathbf{\Theta}$ is a set containing parameters of the model. Optimization methods can be used to find the best set of model parameters, $\hat{\mathbf{\Theta}}$, to minimize the difference between the predicted output $\hat{\mathbf{y}}=f_{\hat{\mathbf{\Theta}}}(\mathbf{x})$ and the ground truth $\mathbf{y}$. 
In regard to the form of $\mathbf{y}$, machine learning problems can be differentiated as classification problems if the value of $\mathbf{y}$ is categorical, or regression problems if the value of $\mathbf{y}$ is continuous.

In the past decades, various machine learning models have been proposed (e.g. support vector machines \citep{cortes1995support}). However, these methods require carefully devised hand-crafted features.
Thanks to the recent growth in computational capacity and the availability of a large amount of training data, deep learning automatically integrates feature extraction and modelling together, and promising performance gains have been observed in many varied tasks. 
Fig. 1 provides an illustration of the difference between conventional machine learning vs. deep learning. 
In the following subsections, the details of deep learning are introduced.

\subsection{Neural Networks}

\begin{figure*}[htbp]
\begin{minipage}[c]{0.49\textwidth}
\centering
\includegraphics[width=\textwidth]{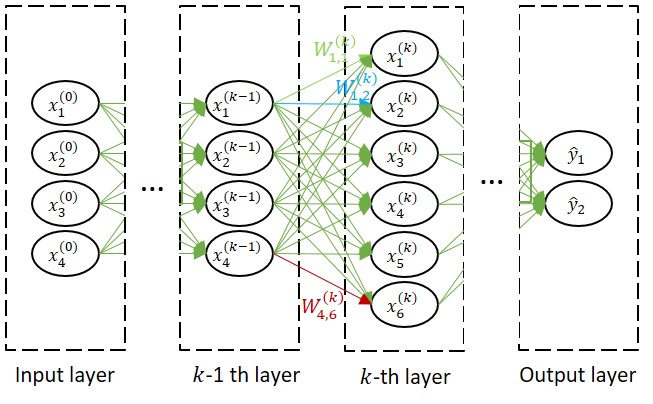}
\caption*{(a)}
\end{minipage}
\begin{minipage}[c]{0.49\textwidth}
\centering
\includegraphics[width=0.8\textwidth]{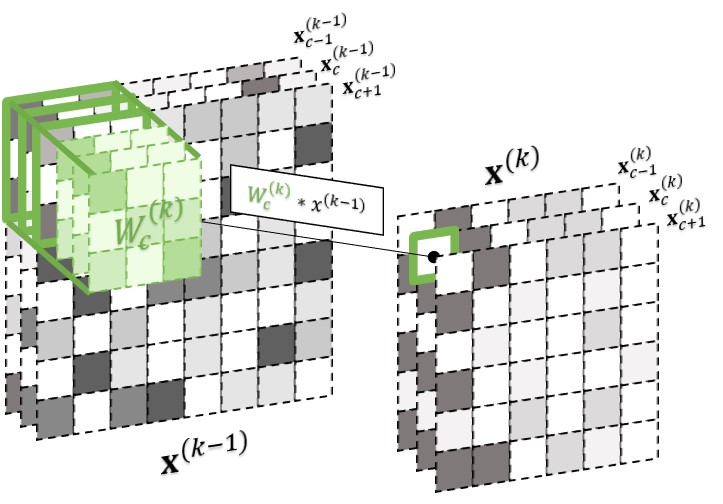}
\caption*{(b)}
\end{minipage}
\caption{(a) Illustration of MLPs with input layer, hidden layers, and output layer. (b) Illustration of the convolution filter $\mathbf{W}^{(k)}_{c}$ that operates with the input $\mathbf{x}^{(k-1)}$ and outputs the $c$-th channel of $\mathbf{x}^{(k)}$.}
\label{fig:mlp}
\end{figure*}

The model $f_{\mathbf{\Theta}}$ in machine learning can be chosen as a neural network (NN) \citep{schmidhuber2015deep}, which contains an interconnected group of nodes (i.e., artificial neurons) inspired and simplified by the biological neural networks in animal brains. 
The most well-known neural network architectures are the multi-layered perceptrons (MLPs), shown in Fig. \ref{fig:mlp} (a). This architecture organizes nodes into groups of layers and connects nodes between neighbouring layers. 
In detail, the computations of the $k$-th layer can be written as:
\begin{equation}
\mathbf{x}^{(k)}=\sigma(\mathbf{W}^{(k)}\mathbf{x}^{(k-1)}+\mathbf{b}^{(k)}),
\label{equ:mlp}
\end{equation}
where $\mathbf{x}^{(k)}\in \mathbf{R}^{p^{(k)}}$ is the input of the $k$-th layer which can be viewed as $p^{(k)}$ nodes of the neural network; $\mathbf{W}^{(k)}\in \mathbf{R}^{p^{(k)}\times p^{(k-1)}}$ with the bias vector $\mathbf{b}^{(k)}\in \mathbf{R}^{p^{(k)}}$ represents a linear transform of the input signal which introduces full connectivity between the $(k-1)$-th layer and the $k$-th layer; $\sigma$ is an activation function which introduces a non-linearity to the output, allowing complex representations. In particular, $\mathbf{x}^{(0)}$ is the input feature 
of a sample in $D$.

The layer defined in Eq. (\ref{equ:mlp}) can also be referred to as a fully connected (FC) layer. By stacking multiple FC layers, neural networks are able to formulate more complex representations of the input. 
To obtain predictions, computations are conducted from the first (input) layer to the last (output) layer, which is known as a forward propagation stage. To optimize the 
parameters $\mathbf{\Theta}=\{\mathbf{W}^{(k)}, \mathbf{b}^{(k)}\}$ of a neural network, a backward propagation stage updates the parameters in an inverse order. 
Recently, more mechanisms and architectures have been proposed for constructing deeper neural networks, of which the ones related to weed recognition are reviewed in the rest of this section.

\subsection{Convolution Neural Networks}

Inspired by the biological processes of animal visual cortex, 
convolution neural networks (CNNs) reduce the challenges of training deep models for visual tasks \citep{gu2018recent}. 
Convolution layers are the key components of CNNs, as illustrated in Fig. \ref{fig:mlp} (b). It involve partial connections compared with fully connected layers of MLPs, where each node focuses on a local region of the input. In detail, denote $W^{(k)} = \{W_{1}^{(k)}, W_{2}^{(k)}, ..., W_{C^{(k)}}^{(k)}\}$ to represent a series of $C^{(k)}$ convolution filters and the computations of the $k$-th convolution layer can be written as:
\begin{equation}
\mathbf{x}^{(k)}_{c}=\sigma(\mathbf{W}^{(k)}_{c}*\mathbf{x}^{(k-1)}+\mathbf{b}^{(k)}),
\label{equ:conv}
\end{equation}
where $*$ represents a convolution operator and $\sigma$ is an activation function; $\mathbf{x}^{(k-1)}$ is the input feature map containing $C^{(k-1)}$ channels and the output feature map is $\mathbf{x}^{(k)} = (\mathbf{x}_{1}^{(k)}, ...,\mathbf{x}_{C^{(k)}}^{(k)})$ containing $C^{(k)}$ channels. 
A convolution layer can be viewed as a special case of FC layers with a sparse weight matrix.

Convolution layers often reduce the input spatial size but increase the number of channels. For some applications, recovery from a deep representation to the original input size is required. 
For this purpose, deconvolution (transpose convolution) operations were used. Readers can refer to \citep{zeiler2010deconvolutional} for more details.

\subsection{Graph Neural Networks}

Whereas most neural networks were designed for processing vectorized data, there are a wide range of applications involving non-vectorized data. Graph neural networks (GNNs) were devised for graph inputs. One commonly used GNN is graph convolution neural networks (GCNNs), which is a generalisation of the conventional CNNs by involving adjacency patterns \citep{bruna2013spectral}. 
In detail, a particular form to compute graph convolutions in $k$-th layer can be written as:
\begin{equation}
\mathbf{X}^{(k)}=\sigma (\mathbf{X}^{(k-1)}A\mathbf{W}^{(k)}+\mathbf{b}^{(k)}),
\label{equ:gconv}
\end{equation}
where $\mathbf{X}^{(k)}\in \mathbf{R}^{n\times p^{(k)}}$ represents the vertex features of $n$ vertices in a graph, $\mathbf{A}\in \mathbf{R}^{n\times n}$ is an adjacency matrix to illustrate the relations between vertices, $\mathbf{W}^{(k)}\in \mathbf{R}^{p^{(k-1)}\times p^{(k)}}$ contains trainable weights and $\mathbf{b}^{(k)}$ is a bias vector. 
Instead of using pre-defined adjacency matrix, graph attention network (GAT) estimated edge weights of the adjacency in line with the vertex features \citep{velickovic2018graph}. 
Recently, various methods were proposed to focus on some graph characteristics, which could not be captured by GCNNs (e.g. longest circle in \citep{garg2020generalization}).

\subsection{Deep Learning Architectures}

Following the above discussed neural networks for deep learning, various deep learning architectures can be constructed in line with different target applications. In terms of weed detection tasks, four categories of deep neural network architectures are summarized, including image classification, object detection, semantic segmentation, and instance segmentation. 

\subsubsection{Image Classification}

Image classification tasks focus on predicting the category (e.g. weed species) of the object in an input image. 
Input images can be viewed as spatially organized data, and many CNN based architectures have been proposed for classifying them into a specific class or category. AlexNet, which consists of 5 convolution layers and 3 fully connected layers was first adopted for large scale image classification~\citep{krizhevsky2012imagenet}. Convolution layers (potentially with other mechanisms) were used to formulate deep representations from input images and FC layers were further used to generate output vectors in line with the categories involved. 
VGG~\citep{simonyan2014very} %, which was invented by the Visual Geometry Group of University of Oxford,
further introduced convolution filters with a $3 \times 3$ perception field to substitute each convolution filter with a large perception field to learn a deeper representation and reduce the computational costs. 
InceptionNet was further proposed to introduce filters of multiple sizes at the same level to characterize the salient regions in an image, which can have extremely large variations in size \citep{szegedy2017inception}.  
With the growing depth of the CNN architectures, short-cuts between the layers alleviate the gradient vanishing issues \citep{8099726}, such as ResNet and DenseNet. NASNet \citep{zoph2018learning} was obtained by architecture search algorithms.

%To further reduce the computational costs of these architectures, methods like MobileNet, SqueezeNet\citep{iandola2016squeezenet} and MNASNet \citep{zoph2018learning} were proposed by involving less parameters while keeping comparable performance. 
%CNNs were also applied to collect temporal patterns from sequentially organized data such as TCN involving the dilated convolution filters \citep{lea2017temporal}. Spatial-temporal data are focused as well and C3D \citep{tran2015learning}, Pseudo 3D (P3D) \citep{qiu2017learning}, R(2+1)D \citep{tran2018closer}, I3D \citep{carreira2017quo}, etc. were proposed for the realms such as video analysis. 

\subsubsection{Object Detection}

Object detection aims to identify the positions and the classes of the objects contained in an image. 
Generally, various CNN architectures for image classification can be used as backbones to learn deep representations and specific output layers can be introduced to obtain object-level annotations including positions and categories. 
%Removing or replacing the FC layers, these classification architectures can be further adopted as backbones for other purposes including the object detection and the semantic segmentation. 
For example, R-CNN and its improvements such as Faster R-CNN~\citep{ren2015faster} were proposed with a two-stage scheme where the first stage generates regional proposals and the second stage predicts the positions and labels for those proposals. One-stage methods were also explored to perform object detection with less latency. For example, YOLO \citep{redmon2016you} (You Only Look Once) was proposed by treating object detection as a regression problem, of which the output is a feature map containing the positions and labels of each pre-defined grid cell. Single 
shot multi-box detector (SSD) \citep{liu2016ssd} introduced feature maps of multiple scales and prior anchor boxes of different ratios. As class imbalance is one of the critical challenges for one-stage object detection, RetinaNet with focal loss was proposed \citep{lin2017focal}. %M2Det \citep{zhao2019m2det} was proposed to explore more effective feature pyramids of different scales for object detection. %A number of improvements on YOLO were studied by exploring multiple additional mechanisms such as Mosaic data augmentation and self-adversarial training in \citep{bochkovskiy2020yolov4}. 
Note that pre-defined anchor boxes played an important role for most of the above mentioned methods. To avoid the significant computational costs to compute such anchor boxes, anchor-free methods were also investigated in \citep{tian2019fcos}. 

\subsubsection{Semantic Segmentation}

Semantic segmentation focuses on the pixel-wise (dense) predictions of an image, by which the category of each pixel is identified. In general, semantic segmentation uses fully convolution networks (FCNs), which were first explored in \citep{long2015fully}. 
These studies often involved an encoder-decoder scheme: the encoder formulates a latent representation of an image through convolutions and the decoder focuses on upsampling of the latent representation to the original image size for dense predictions. By increasing the capacity of the decoder, U-Net \citep{ronneberger2015u} was proposed with promising performance for medical images. 
SegNet \citep{badrinarayanan2017segnet} additionally involved pooling indices for its decoder, compared with the general encoder-decoder models to use the pooled values only, to perform non-linear upsampling to keep the boundary information. Pyramid scene parsing network (PSPNet) and  different-region-based context aggregation by a pyramid pooling module \citep{zhao2017pyramid} exploited the capability of global context as a superior framework for pixel-level predictions. 
Instead of following a conventional encoder-decoder scheme, DeepLab models adopted atrous convolutions to reduce the downsampling operations with a large reception field \citep{chen2018encoder}. %In recent, further improvements were made in regard to segmentation quality and efficiency, such as DeepLabv3+ \citep{chen2018encoder} and FastFCN \citep{wu2019fastfcn}.

\subsubsection{Instance Segmentation}
Instance segmentation aims to output both the class and class instance information for individual pixels. Instance segmentation methods were initially devised in a two-stage manner by focusing on two separated tasks: object detection and semantic segmentation. For example, Mask R-CNN~\citep{he2017mask} was proposed using a top-down design, which first conducts object detection task to locate the bounding boxes of each instance and next within each bounding box undertakes semantic segmentation. Bottom-up methods were also investigated, which first conduct semantic segmentation and use clustering or metric learning to obtain different instances (e.g.~\citep{papandreou2018personlab}). Two-stage methods require accurate results from each stage %, otherwise, for example, the performance in terms of instance edges could be dropped significantly. In addition, 
and the computation cost of two-stage methods could be expensive. 
Therefore, single shot methods were explored. Anchor-based methods YOLACT~\citep{bolya2019yolact} introduced two parallelized tasks to an existing one-stage object detection model including a dictionary of non-local prototype masks over the entire image and predicting a set of linear combination coefficients per instance. 
%and SOLO~\citep{wang2019solo} were studied. 
An anchor-free method fully convolutional instance-aware semantic segmentation (FCIS) was proposed based on FCNs by introducing position-sensitive inside/outside score maps % to output semantic segmentation and instance masks simultaneously 
\citep{li2017fully}. 
PolarMask~\citep{xie2020polarmask} conducted instance center classification and dense distance regression in a polar coordinate, which is a much simpler and flexible framework.  
Very recently, BlendMask~\citep{chen2020blendmask} inspired by FCIS introduced a blender module to effectively combine instance-level information and semantic information with low-level fine-granularity. 

\section{Deep Learning for Weed Recognition}
\label{sec:weed}

In this section, deep learning based weed recognition studies are summarized covering four 4 approaches: image classification, object detection, semantic segmentation and instance segmentation. Before reviewing those approaches, research data, data augmentation and evaluation metrics used in these studies are reviewed first to provide a context in the field. %In particular, to facilitate future research, various public datasets for weed management are presented. 

\subsection{Weed Data}
\label{sec:dataset}

Weed data is the foundation for developing and benchmarking weed recognition methods, and sensing technologies will determine what weed data can be acquired and what weed management practice can be developed~\citep{machleb2020sensor}.
While various sensing techniques like ultrasound
%\citep{Andujar2011}
, LiDAR (Light Detection And Ranging) and optoelectronic sensors 
% \citep{Andujar2013}
%\citep{Peteinatos2014} 
were used for simple differentiation between weeds and crops, image-based weed recognition has gained increasing interests due to the advances of various imaging techniques. %devices and its convenience and promise for precise identification. 
%The simple detection of plants, with limited differentiation capability is possible through the interpretation of data from a variety of sources, including optoelectronic sensors \citep{Peteinatos2014}, ultrasound \citep{Andujar2011} and LiDAR (Light Detection And Ranging) \citep{Andujar2013}. However, image-based data provided the additional information required for crop-weed and weed species differentiation. Different imaging technologies can be adopted for weed identification methods and platforms such as multispectral camers, RGB (Red Green Blue) cameras and RGB cameras with a depth sensor. 

Multispectral imaging captures light energy within specific wavelength ranges or bands of the electromagnetic spectrum, 
%The used spectral regions are often at least partially outside the visible light range (infrared and ultra-violet).  
which can capture information beyond visible wavelength \citep{farooq2018analysis}. % with its visible receptors for Red, Green and Blue. 
For example, hyperspectral images consist of many contiguous and narrow bands;  near infrared (NIR) imaging uses a subset of the infrared band, as the pigment in plant leaves, chlorophyll, strongly absorbs red and blue visible light and reflects near infrared. 
 %Therefore, these images provide detailed spectral characteristics within and beyond visible wavelength for weed identification \citep{farooq2018analysis}. 
%\citep{AgroAVNET for crops and weeds classification: A step forward in automatic farming }
%RGB images are widely used by researchers in remote sensing community. This is because the RGB images are simple to capture and easy to use. Several patch- based methods have been proposed using the RGB images for the classification of crops and weeds. 
Driven by the low cost RGB cameras and the significant progresses in computer vision, %Recently, the use of low-cost RGB cameras, capturing visible light for imaging, 
RGB images have been increasingly used (e.g.~\citep{olsen2019deepweeds}). 
%For example, deep learning techniques are able to achieve promising weed identification performance without manually designing hand-crafted features.
%can be helpful to obtain adequate information from RGB images alone compared with the conventional machine learning methods using carefully devised hand-crafted features. The RGB imaging avoids the added computational and monetary expense associated with multi- and hyperspectral imaging devices. 
\kun{In addition, some studies involved the fusion of depth (the distance between the image plane and each pixel) and RGB images using sensors such as the Kinect v2. The result improved segmentation from 76.4\% for colour-only, to  96.6\% for broccoli \citep{gai2020automated}. }
%, which is used to collect depth information - the distance between the image plane and each pixel - as a new channel
%\citep{brilhador2019classification,wang2020semantic}. 

\begin{figure*}[!htb]
\centering
\includegraphics[width=0.95\textwidth]{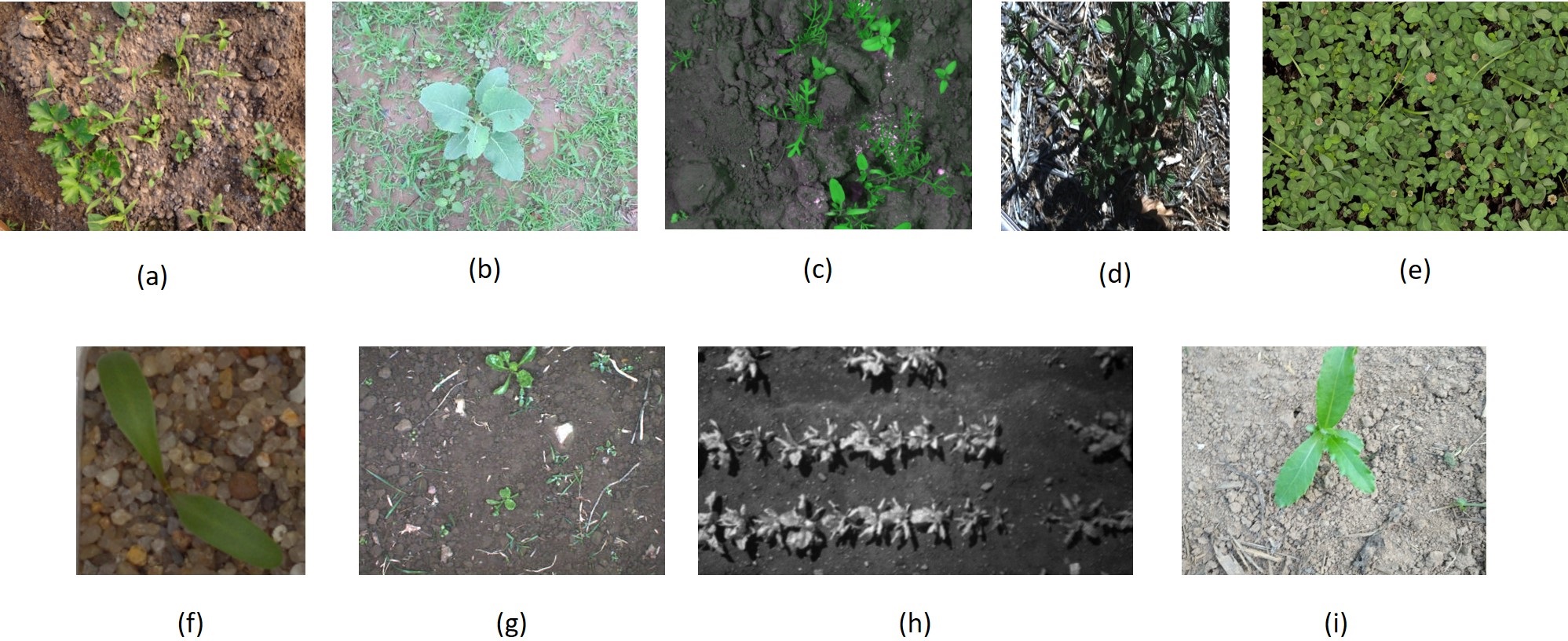}
\caption{Sample weed images from several public datasets. (a) Carrot-Weed \citep{lameski2017weed} (b) CWF-788 \citep{li2019real} (c) CWF-ID \citep{haug2014crop}  (d) DeepWeeds \citep{olsen2019deepweeds} (e) GrassClover \citep{skovsen2019grassclover} (f) Plant Seedlings Dataset \citep{giselsson2017public} (g) Sugar Beets 2016 \citep{chebrolu2017agricultural} (h) Sugar Beet/Weed Dataset \citep{sa2017weednet} (i) Weed-Corn/Lettuce/Radish \citep{jiang2020cnn}.}
\label{fig:dataset_samples}
\end{figure*}

The availability of rich public datasets in the field plays a key role in facilitating the development of new algorithms specific to weed recognition tasks. % and in the fair performance evaluation on the state-of-the-art algorithms. 
%In recent years, a number of datasets specific to crop and weed differentiation have been publicly available. These datasets are summarized below including details on data modality, size, classes and URLs if available. 
In recent years, a number of in-crop weed datasets have been made public, as shown in Fig. \ref{fig:dataset_samples}, and will be reviewed in the rest of this section. % shows sample images of these datasets.

\textbf{Bccr-segset} \citep{NguyenThanhLe2019} contains 30,000 RGB images with pixel-wise annotations collected of canola (\textit{Brassica napus}), maize (\textit{Zea mays}) and wild radish (\textit{Raphanus raphanistrum}). The images were acquired using an gantry-system mounted above an indoor growth facility across multiple growth stages. %This dataset provided pixel-level annotations. 

\textbf{{Carrot-Weed}} \citep{lameski2017weed} contains 39 RGB images collected with a 10-MP (Mega Pixel) phone camera under variable light conditions of young carrot (\textit{Daucus carota subsp. sativus}) seedlings in the Republic of Macedonia. Pixel-level annotations were provided of three categories: carrots, unspecified weeds and soil (\url{https://github.com/lameski/rgbweeddetection}). 

\textbf{{Crop/Weed Field Image Dataset}} (CWFID) \citep{haug2014crop} comprises 60 top-down field images of carrots with intra-row and close-to-crop weeds captured by RGB cameras. Pixel-level annotations are provided for crop vs weed discrimination of 162 carrot plants and 332 weeds in total (\url{https://github.com/cwfid}). 

\textbf{{CWF-788}} \citep{li2019real} is a field image dataset containing 788 RGB images collected from cauliflower (\textit{Brassica oleracea var. botrytis}) fields with high weed pressure. It was collected for semantic segmentation of the cauliflower plants from the background (combining both weeds and soil) with manually segmented annotations (\url{https://github.com/ZhangXG001/Real-Time-Crop-Recognition}).

\textbf{DeepWeeds} \citep{olsen2019deepweeds} was collected from remote rangelands in northern Australia for weed specific image classification. It includes 17,509 images of 8 types of target weed species with various off-target plants native to Australia. The target weed species include chinee apple (\textit{Ziziphus mauritiana}), lantana (\textit{Lantana camara}), parkinsonia (\textit{Parkinsonia aculeata}), parthenium (\textit{Parthenium hysterophorus}), prickly acacia (\textit{Vachellia nilotica}), rubber vine (\textit{Cryptostegia grandiflora}), siam weed (\textit{Chromolaena odorata}) and snake weed (\textit{Stachytarpheta} spp.). 
For each target weed species (positive class), around 1,000 images were obtained; off-target flora and backgrounds not containing the weeds of interest are collected as a single negative class of 9,106 images. It was collected from 8 different locations with an attempt to balance scene bias images of the target species, and negative cases were collected at each location in similar quantities (\url{https://github.com/AlexOlsen/DeepWeeds}).

\textbf{Grass-Broadleaf} \citep{dyrmann2016plant} contains 22 different plant species at early growth stages, which was constructed by combining 6 image datasets. In total, 10,413 RGB images were included. Note that image background was removed in this dataset and each image only contains one individual plant. 

\textbf{{GrassClover}} \citep{skovsen2019grassclover} is a diverse image and biomass dataset, of which 8,000 synthetic RGB images are provided with pixel-wise annotations for semantic segmentation based weed recognition studies. The dataset was collected in an outdoor field setting including 6 classes: unspecified grass species, white clover (\textit{Trifolium repens}), red clover (\textit{Trifolium pratense}), shepherd's purse (\textit{Capsella bursa-pastoris}), unspecified thistle, dandelion (\textit{Taraxacum} spp.) and soil. In addition, 31,600 unlabelled images were provided for pre-training, weakly-supervised learning, and unsupervised learning (\url{https://vision.eng.au.dk/grass-clover-dataset}). 

\textbf{{Plant Seedling Dataset}} \citep{giselsson2017public} contains 960 unique plants at several growth stages in RGB images for species including blackgrass (\textit{Alopecurus myosuroides}), charlock (\textit{Sinapis arvensis}), cleavers (\textit{Galium aparine}), common chickweed (\textit{Stellaria media}), wheat, fat hen (\textit{Chenopodium album}), loose silky-bent (\textit{Apera spica-venti}), maize (\textit{Zea mays}), scentless mayweed, shepherd's purse, small-flowered cranesbill (\textit{Geranium pusillum}) and sugar beet (\textit{Beta vulgaris var. altissima}) (\url{https://vision.eng.au.dk/plant-seedlings-dataset}). 

\textbf{\href{{https://data.mendeley.com/datasets/3fmjm7ncc6/2}}{Soybean/Grass/Broadleaf/Soil}} \citep{dos2017weed} comprises 15,336 segments of soybean (\textit{Glycine max}), unspecified grass weeds, unspecified broadleaf weeds and soil. The segments were extracted using the simple linear iterative clustering (SLIC) superpixel algorithm on 400 images collected with an Unmanned Aerial Vehicle (UAV)-mounted RGB camera. 

\textbf{{Sugar Beets 2016}} \citep{chebrolu2017agricultural} was collected from agricultural fields with pixel-wise annotations for three classes: sugar beet, weeds, and soil. This dataset contains 1,600 images, of which 700 images were captured at first and 900 images were captured after a four-week period. Both RGB-D and multispectral images were provided, which is helpful to explore the effectiveness of different modalities for weed recognition and to construct multi-modal learning methods (\url{http://www.ipb.uni-bonn.de/data/sugarbeets2016}). 

\textbf{{Sugar Beet/Weed Dataset}} \citep{sa2017weednet} contains 155 multispectral images (near-infared 790 nm, red 660 nm) plus normalised difference vegetation index (NDVI) with pixel-wise labelling for sugar beet, weeds and soil from a controlled field experiment (\url{https://github.com/inkyusa/weedNet}). 

\kun{\textbf{{Weed-AI}} is an open-source weed dataset upload and download platform that standardises metadata reporting with the WeedCOCO annotation format and centralises weed datasets. An annotation platform built on the computer vision annotation tool (CVAT) has been integrated. Weed-AI currently contains 17 datasets (including DeepWeeds) with 20,891 images.   
(\url{https://weed-ai.sydney.edu.au/about}). }

\textbf{{Weed-Corn/Lettuce/Radish}} \citep{jiang2020cnn} contains 7,200 RGB images with image-level annotations. It includes three subsets: the maize dataset (1,200 images) was collected from a corn field with four different weed species (4,800 images) including Canada thistle (\textit{Cirsium arvense}), fat hen, bluegrass (\textit{Poa} spp.) and sedge (\textit{Carex} spp.); the lettuce dataset was collected from a vegetable field of two plant classes including lettuce (500 images) and weeds (300 images); the radish dataset contains 200 radish images and 200 weeds images~\citep{lameski2017weed}
(\url{https://github.com/zhangchuanyin/weed-datasets}).

Whilst these datasets provide useful imagery and annotation data for benchmarking, there is a lack of consistency and details in metadata reporting standards and contextual information. 
An understanding of weed species, beyond a simple awareness of the difference to crops, is important in creating opportunities to deliver specific weed control treatments. % to different weed species. 
%As weed species identification becomes increasingly important for targeted applications, it is essential that consistency in species reporting is maintained. 
For example, contextual understanding of crop growth stages, presence/absence of stubble will assist in developing algorithms capable of handling variability across different conditions.

\subsection{Data Augmentation}

Due to the laborious nature of developing annotated datasets within weed control contexts, existing datasets are often not large enough and do not reflect sufficient diversity in conditions. 
A significant risk for deep learning using small datasets is overfitting, where the model performs well on the training set but performs poorly %on test datasets or 
when being deployed in the fields. % with different conditions. 
To address this issue, %existing deep learning studies including those for weed identification have adopted 
various data augmentation strategies were adopted to %artificially 
enlarge the size and the quality of the training sets such as random cropping, rotation, flipping, color space transformation, noise injection, image mixing, random erasing and generative approaches. %These techniques are helpful for weed identification. 
Readers can refer to \citep{shorten2019survey} for more details.

\subsection{Evaluation Metrics}
A number of metrics have been utilised to evaluate the desktop performance of weed recognition algorithms. The definition of these metrics may differ in terms of different types of recognition approaches. \kun{The focus on desktop-based evaluation metrics instead of real-world field evaluation metrics is seen as a short coming of current methods \citep{salazar2021towards}. Nevertheless, the metrics are the current standard for comparison.} %Developing consistent understandings of real-world implications of the meanings of these metrics is critical in translating research for weed identification into real-world outcomes.

For binary image classification whereby the classification result of an input sample is labelled either as a positive (P) or a negative (N) case, 4 possible outcomes can be derived:
%The building blocks of these metrics are first introduced including the concepts of \textbf{true positive} (TP), \textbf{false positive} (FP), \textbf{true negative} (TN) and \textbf{false negative} (FN). 
(1) If a positive sample is classified as positive, the prediction is correct and defined as \textbf{true positive} (TP). (2) If a negative sample is classified as positive, the prediction is \textbf{false positive} (FP). (3) If a negative sample is classified as negative, the prediction is \textbf{true negative} (TN). (4) If a positive sample is classified as negative, the prediction is \textbf{false negative} (FN). %Fig. \ref{fig:metrics} illustrates these definitions visually. %Note that the concept of a sample varies with the approach used. 
%\begin{figure}[!htb]
%\centering
%\includegraphics[width=0.45\textwidth]{metrics.jpg}
%\caption{Illustration of the true positives (TP), false negatives (FN), true negatives (TN), false positives (FP).}
%\label{fig:metrics}
%\end{figure}

Based on these definitions, some widely used evaluation metrics can be defined for benchmarking the performance of different algorithms. 
%It is important to note that algorithm accuracy, for example, on a test dataset and accuracy when measured in the field on a per-plant basis may different significantly.
\textbf{Accuracy} measures the proportion of the correct predictions (\#TP + \#TN)
%(including both the true positive and true negative) 
over all the predictions (\#P + \#N).
%\begin{equation}
%Acc. = \frac{\#TP + \#TN}{\#P + \#N}.
%\label{equ:acc}
%\end{equation}
\textbf{Sensitivity}, also known as \textbf{recall}, measures the proportion of the correctly predicted positive cases (\#TP) over all the positive cases (\#TP + \#FN). 
%\begin{equation}
%Sens. (Recall) = \frac{\#TP}{\#TP + \#FN}.
%\label{equ:sens}
%\end{equation}
%For weed detection, it represents the ratio between the total number of correct weed detection cases and the total number of weeds available, 
It indicates the likelihood that the algorithm identifies all weeds. A low sensitivity would suggest that a large number of weeds are missed, while a sensitivity rate 1 indicates that all weeds are successfully detected. %The computation of sensitivity can be written as:
\textbf{Precision} measures the proportion of the correctly predicted positive cases (\#TP) over all predicted positive cases (\#TP + \#FP). 
%\begin{equation}
%Prec. = \frac{\#TP}{\#TP + \#FP}.
%\label{equ:precision}
%\end{equation}
For weed detection, a high precision indicates low off-target or crop damage. %while a low precision means that more crops would be unnecessarily treated. 
\textbf{Specificity} measures the proportion of the correctly predicted negative cases (\#TN) over all predicted negative cases (\#TN + \#FP).
%\begin{equation}
%Spec. = \frac{\#TN}{\#TN + \#FP}.
%\label{equ:spec}
%\end{equation}
A low specificity suggests that an algorithm is delivering an control option towards crops. 
%The values of all these metrics are within $[0,1]$ and expected to be as close to 1 as possible for promising weed detection methods. %There is as yet little research comparing the in-field performance of deep learning algorithms on a known number of weeds with the performance on a test dataset. 
%For example, if there are no false negative cases, the precision is 1 according to Eq. (\ref{equ:precision}). 
\textbf{F-score} (also known as \textbf{F$_{1}$ score}) combines the precision and the recall values by treating them with equal importance: 
\begin{equation}
F_{1} = 2\cdot\frac{Precision\cdot Recall}{Precision + Recall}.
\label{equ:precision}
\end{equation}

As a binary classification model generally outputs continuous predictions, a threshold is required to judge the predicted labels: if the score is beyond the threshold, the corresponding sample is predicted as a positive case; otherwise the sample is predicted as a negative case. By varying the threshold, %the above mentioned metrics can be changed as well. Hence, 
trade-offs among some metrics %according to the applications 
can be made. %, especially for the sensitivity and specificity trade-off and the precision and recall trade-off. 
%Two types of curves are often adopted to illustrate the ability of a binary model for the trade-off purposes. 
A receiver operating characteristic curve (\textbf{ROC curve}) illustrates the diagnostic ability of a binary classification model by plotting the sensitivity against the 1-specificity at various threshold settings. A precision-recall curve (\textbf{PR curve}) shows plots the precision against the recall. A large area under these curves (\textbf{AUC}) often indicates a model of high quality. For multi-class classification, most of the these metrics can be computed class by class and the mean of these metrics can be used.% have been used for evaluation in existing weed management studies.% (e.g.~\citep{olsen2019deepweeds}). 

For an object detection task with only one class, a sample is associated with an object in a bounding box. For a predicted bounding box, \textbf{intersection over union} (IoU) is defined as the area of the intersection divided by the area of the union of the predicted bounding box and a ground truth bounding box. % as in Fig. 5. 
%\begin{figure*}[htbp]
%\begin{minipage}[c]{0.55\textwidth}
%\centering
%\includegraphics[width=\textwidth]{iou.jpg}
%\end{minipage}
%\begin{minipage}[c]{0.43\textwidth}
%\caption{Illustration of the computations of IoU, where the dark green bounding box is the ground truth and the light green bounding box is the prediction result.}
%\end{minipage}
%\label{fig:ioux}
%\end{figure*}
%\begin{figure}[!htb]
%\centering
%\includegraphics[width=0.45\textwidth]{iou.jpg}
%\caption{Illustration of the computations of IoU, where the dark green bounding box is the ground truth and the light green bounding box is the prediction result.}
%\label{fig:iou}
%\end{figure}
%\begin{equation}
%IoU = \frac{area\;of\;overlap}{area\;of\;union}.
%\label{equ:precision}
%\end{equation}
Given a threshold $\theta$, if the confidence value of a predicted bounding box is beyond $\theta$ and the IoU against the ground truth bounding box is beyond 0.5, the predicted bounding box is regarded as a TP case; if the confidence is beyond $\theta$ and the IoU is less than 0.5, it is regarded as a FP case; if the confidence is less than $\theta$ and the IoU is less than 0.5, it is regarded as a TN case; if the confidence is less than $\theta$ and the IoU is beyond 0.5, it is regarded as a FN case. 
Next, the precision and recall values can be defined to measure the quality of detection results. By varying $\theta$, a PR curve can be obtained and \textbf{average precision} (AP): $\int_{0}^{1} p(r){\rm d}r$ is used to summarize the quality of the PR curve,
%\begin{equation}
%AP = \int_{0}^{1} p(r){\rm d}r,
%\label{equ:ap}
%\end{equation}
where $p(r)$ indicates the precision value $p$ corresponding to the recall values $r$ for a particular IoU threshold. In practice, different estimations for AP are adopted such as the AUC of the PR curve. 
Different IoU threshold values other than 0.5 can also be used to select the bounding boxes from the candidates and the corresponding AP can be obtained. For example, AP$_{50}$ and AP$_{75}$ define the AP with IoU threshold 0.5 and 0.75, respectively.  
For multi-class object detection problems, these metrics can be computed for each class individually and a \textbf{mean average precision} (mAP) can be obtained over all classes. 

In segmentation tasks, a sample can be viewed as a pixel. The metrics such as (mean) accuracy (mAcc), recall, precision and F-score discussed above can be derived in a similar manner. By organizing the pixels belonging to the same class as regions, the concepts such as mAP and mIoU can be derived as well. 

% Table generated by Excel2LaTeX from sheet 'Sheet7'
\begin{table}[htbp]
  \centering
  \caption{\kun{Comparisons between different weed recognition methods}}
  \begin{threeparttable}
  \scalebox{0.75}{
    \begin{tabular}{l|ccc|ccc|ccccccc}
    Approaches & \multicolumn{3}{c|}{Image Classification} & \multicolumn{3}{c|}{Object Detection} & \multicolumn{4}{c|}{Semantic Segmentation} & \multicolumn{3}{c}{Instance Segmentation} \\
    \hline
    Granularity & \multicolumn{3}{c|}{Image-level} & \multicolumn{3}{c|}{Plant-level} & \multicolumn{4}{c|}{Pixel-level} & \multicolumn{3}{c}{Pixel and plant-level} \\
    \tabincell{l}{Precision} & \multicolumn{3}{c|}{Low} & \multicolumn{3}{c|}{Medium} & \multicolumn{4}{c|}{High}      & \multicolumn{3}{c}{Very high} \\
    \tabincell{l}{Annotation\\Intensity} & \multicolumn{3}{c|}{Low} & \multicolumn{3}{c|}{Medium} & \multicolumn{4}{c|}{High}      & \multicolumn{3}{c}{High} \\
    \cellcolor[HTML]{C0C0C0}\tabincell{l}{\textit{Example}}      &       &  \multicolumn{1}{c}{\cellcolor[HTML]{C0C0C0}ResNet50} & \multicolumn{1}{c|}{\cellcolor[HTML]{C0C0C0}Swin-T} &       & \multicolumn{1}{c}{\cellcolor[HTML]{C0C0C0}YOLOv6-S} & \multicolumn{1}{c|}{\cellcolor[HTML]{C0C0C0}YOLOv6-M} &       & \multicolumn{1}{c}{\tabincell{c}{\cellcolor[HTML]{C0C0C0}DeepLabV3\\\cellcolor[HTML]{C0C0C0}\\\cellcolor[HTML]{C0C0C0}(ResNet50)}} & \multicolumn{1}{c}{\tabincell{c}{\cellcolor[HTML]{C0C0C0}DeepLabV3\\\cellcolor[HTML]{C0C0C0}\\\cellcolor[HTML]{C0C0C0}(ResNest200)}} & \multicolumn{1}{c|}{\tabincell{c}{\cellcolor[HTML]{C0C0C0}UpperNet\\\cellcolor[HTML]{C0C0C0}\\\cellcolor[HTML]{C0C0C0}(Swin-L)}} &       & \multicolumn{1}{c}{\tabincell{c}{\cellcolor[HTML]{C0C0C0}MaskRCNN\\\cellcolor[HTML]{C0C0C0}\\\cellcolor[HTML]{C0C0C0}(ResNet50)}} & \multicolumn{1}{c}{\tabincell{c}{\cellcolor[HTML]{C0C0C0}Cascade\\\cellcolor[HTML]{C0C0C0}MaskRCNN\\\cellcolor[HTML]{C0C0C0}(Swin-T)}} \\
    FLOPS & \multicolumn{1}{l}{Low} & \multicolumn{1}{c}{\cellcolor[HTML]{C0C0C0}4G} & \multicolumn{1}{c|}{\cellcolor[HTML]{C0C0C0}4.5G} & \multicolumn{1}{l}{Medium} & \multicolumn{1}{c}{\cellcolor[HTML]{C0C0C0}44.2G} & \multicolumn{1}{c|}{\cellcolor[HTML]{C0C0C0}82.2G} & \multicolumn{1}{l}{High} & \multicolumn{1}{c}{\cellcolor[HTML]{C0C0C0}51.4G} & \multicolumn{1}{c}{\cellcolor[HTML]{C0C0C0}1381G} & \multicolumn{1}{c|}{\cellcolor[HTML]{C0C0C0}3230G} & \multicolumn{1}{l}{High} & \multicolumn{1}{c}{\cellcolor[HTML]{C0C0C0}260G} & \multicolumn{1}{c}{\cellcolor[HTML]{C0C0C0}745G} \\
    \tabincell{l}{Model Size\\(\# Params)} & \multicolumn{1}{l}{Small} & \multicolumn{1}{c}{\cellcolor[HTML]{C0C0C0}25.6M} & \multicolumn{1}{c|}{\cellcolor[HTML]{C0C0C0}29M} & \multicolumn{1}{l}{Medium} & \multicolumn{1}{c}{\cellcolor[HTML]{C0C0C0}17.2M} & \multicolumn{1}{c|}{\cellcolor[HTML]{C0C0C0}34.3M} & \multicolumn{1}{l}{High} & \multicolumn{1}{c}{\cellcolor[HTML]{C0C0C0}42.0M} & \multicolumn{1}{c}{\cellcolor[HTML]{C0C0C0}88M} & \multicolumn{1}{c|}{\cellcolor[HTML]{C0C0C0}234M} & \multicolumn{1}{l}{High} & \multicolumn{1}{c}{\cellcolor[HTML]{C0C0C0}46.4M} & \multicolumn{1}{c}{\cellcolor[HTML]{C0C0C0}86M} \\
    \tabincell{l}{Power\\Consumption\tnote{*}} & \multicolumn{1}{l}{Low} & \multicolumn{1}{c}{\cellcolor[HTML]{C0C0C0}0.08W} & \multicolumn{1}{c|}{\cellcolor[HTML]{C0C0C0}0.09W} & \multicolumn{1}{l}{Medium} & \multicolumn{1}{c}{\cellcolor[HTML]{C0C0C0}0.90W} & \multicolumn{1}{c|}{\cellcolor[HTML]{C0C0C0}1.68W} & \multicolumn{1}{l}{High} & \multicolumn{1}{c}{\cellcolor[HTML]{C0C0C0}1.05W} & \multicolumn{1}{c}{\cellcolor[HTML]{C0C0C0}28.18W} & \multicolumn{1}{c|}{\cellcolor[HTML]{C0C0C0}65.92W} & \multicolumn{1}{c}{High} & \multicolumn{1}{c}{\cellcolor[HTML]{C0C0C0}5.31W} & \multicolumn{1}{c}{\cellcolor[HTML]{C0C0C0}15.20W} \\
    \tabincell{l}{Training \&\\Inference Speed} & \multicolumn{3}{c|}{Fast} & \multicolumn{3}{c|}{Medium} & \multicolumn{4}{c|}{Slow}      & \multicolumn{3}{c}{Slow} \\
    Scenarios & \multicolumn{3}{c|}{\tabincell{c}{Large whole plants}} & \multicolumn{3}{c|}{\tabincell{c}{Whole-plant, leaf, plant organ}} & \multicolumn{4}{c|}{Whole-plant, leaf, plant organ} & \multicolumn{3}{c}{Whole-plant, leaf, plant organ}  \\
    \cellcolor[HTML]{C0C0C0}\textit{Example} & \multicolumn{3}{c|}{\cellcolor[HTML]{C0C0C0}\citep{zhuang2022evaluation}} & \multicolumn{3}{c|}{\cellcolor[HTML]{C0C0C0}\citep{sharpe2019detection}} & \multicolumn{4}{c|}{\cellcolor[HTML]{C0C0C0}\citep{picon2022deep}} & \multicolumn{3}{c}{\cellcolor[HTML]{C0C0C0}\citep{champ2020instance}} \\
    \tabincell{l}{Weed Control\\Treatment} & \multicolumn{3}{c|}{Spot spraying} & \multicolumn{3}{c|}{Spot spraying} & \multicolumn{4}{c|}{Precision spraying/laser weeding/tillage} & \multicolumn{3}{c}{Laser weeding} \\
    \cellcolor[HTML]{C0C0C0}\textit{Example} & \multicolumn{3}{c|}{\cellcolor[HTML]{C0C0C0}\citep{calvert2021robotic}} & \multicolumn{3}{c|}{\cellcolor[HTML]{C0C0C0}\citep{salazar2021towards}} & \multicolumn{4}{c|}{\cellcolor[HTML]{C0C0C0}\citep{bawden2017robot}} & \multicolumn{3}{c}{\cellcolor[HTML]{C0C0C0}-} \\
    \end{tabular}%
    }
    
    \begin{tablenotes}
        \footnotesize
        \item[*] Estimated with an Nvidia GTX 1080Ti GPU. 
    \end{tablenotes}
    \end{threeparttable}
  \label{tab:tb_approaches}%
\end{table}%

\begin{table*}[htbp]
  \centering
  \caption{Summary  of image classification based weed recognition studies.}
  \begin{threeparttable}
  \scalebox{0.75}{
    \begin{tabular}{llllllrllll}
    \toprule
    Architecture & \tabincell{l}{Type} & Dataset & \tabincell{l}{Public\\ Avail.} & \tabincell{l}{Target\\ Species} & AUC   & \multicolumn{1}{l}{Acc.} & Prec. & \tabincell{l}{Recall\\ Sens.} & \tabincell{l}{$F_1$\\ Score} & Spec. \\
    \midrule
    \multicolumn{11}{c}{\textbf{Hyperspectral (400 - 1000nm)}}\\
    \midrule
    \tabincell{l}{Customized \\CNN (500X500) \citep{farooq2018analysis}} & Multiclass & \multirow{4}*{\tabincell{l}{\\ \\ UNSW\\ Hyperspectral\\ Weed Dataset}} & Y     &  4 weeds & -     & 94.7  & -     & -     & -     & - \\
    \tabincell{l}{Customized \\CNN (250X250) \citep{farooq2018analysis}} & Multiclass & ~ & Y     &  4 weeds & -     & 90.6  & -     & -     & -     & - \\
    \tabincell{l}{Customized \\CNN (125X125) \citep{farooq2018analysis}} & Multiclass & ~ & Y     &  4 weeds & -     & 86.3  & -     & -     & -     & - \\
    \tabincell{l}{FCNN-SPLBP \\ (100X100) \citep{farooq2019multi}} & Multiclass & ~ & Y & 4 weeds & -     & 89.7  & -     & -     & -     & - \\
    \midrule
    \multicolumn{11}{c}{\textbf{Multispectral (NIR \& RGB)}}\\
    \midrule
    FCNN-SPLBP \citep{farooq2019multi} & Multiclass & \tabincell{l}{Sugar beet/weed\\ Dataset}  & Y     &  \tabincell{l}{crops, weeds,\\ mix} & -     & 96.4  & -     & -     & -     & - \\
    \midrule
    \multicolumn{11}{c}{\textbf{RGB}}\\
    \midrule
    ResNet-18 \citep{bah2018deep} & Binary & Bean  & -      & bean, weeds & \multicolumn{1}{r}{94.8} & \multicolumn{1}{l}{-} & -     & -     & -     & - \\
    \midrule
    ResNet-18 \citep{bah2018deep} & Binary & Spinach & -      & spinach, weeds & \multicolumn{1}{r}{95.7} & \multicolumn{1}{l}{-} & -     & -     & -     & - \\
    \midrule
    AgroAVNET \citep{chavan2018agroavnet} & Multiclass & \tabincell{l}{Plant Seedling\\ Dataset} & Y       & \tabincell{l}{3 crops,\\9 weeds} & -     & 93.6  & \multicolumn{1}{r}{93.0} & \multicolumn{1}{r}{94.0} & \multicolumn{1}{r}{93.0} & - \\
    \midrule
    GCN-ResNet-101 \citep{jiang2020cnn} & Binary & Radish & Y       & radish, weeds & -     & 98.9  & \multicolumn{1}{r}{98.5} & \multicolumn{1}{r}{98.3} & \multicolumn{1}{r}{98.9} & \multicolumn{1}{r}{98.5} \\
    \midrule
    GCN-ResNet-101 \citep{jiang2020cnn} & Binary & Corn  & Y       & corn, weeds & -     & 97.8  & \multicolumn{1}{r}{99.3} & \multicolumn{1}{r}{99.2} & \multicolumn{1}{r}{99.3} & \multicolumn{1}{r}{97.1} \\
    \midrule
    GCN-ResNet-101 \citep{jiang2020cnn} & Binary & Lettuce & Y       & lettuce, weeds & -     & 99.4  & \multicolumn{1}{r}{98.4} & \multicolumn{1}{r}{99.7} & \multicolumn{1}{r}{99.0} & \multicolumn{1}{r}{99.0} \\
    \midrule
    GCN-ResNet-101 \citep{jiang2020cnn} & Binary & Mixed Dataset & Y       & 3 crops, weeds & -     & 96.5  & \multicolumn{1}{r}{98.8} & \multicolumn{1}{r}{98.7} & \multicolumn{1}{r}{97.2} & \multicolumn{1}{r}{96.5} \\
    \midrule
    ResNet-50 \citep{olsen2019deepweeds} & Multiclass\tnote{*} & \multirow{4}*{\tabincell{l}{\\DeepWeeds}} & Y      & 8 weeds, others & -     & 95.7  & \multicolumn{1}{r}{95.7} & -     & -     & \multicolumn{1}{r}{98.0} \\
    DenseNet-128-32 \citep{lammie2019low} & Multiclass\tnote{*} & ~ & Y      & 8 weeds, others & -     & 90.1  & -     & -     & -     & - \\
    GraphWeedsNet \citep{hu2020graph} & Multiclass\tnote{*} & ~ & Y      & 8 weeds, others & -     & 98.1  & \multicolumn{1}{r}{98.2} & -     & -     & \multicolumn{1}{r}{99.3} \\
    \tabincell{l}{DeepCluster \\VGG-16 \citep{dyrmann2016plant}}  & Multiclass & ~ & Y      & 8 weeds, others & -     & 70.6  & \multicolumn{1}{r}{96.6} & -     & -     & - \\
    \midrule
    \tabincell{l}{Customized \\Residual Net\\(128X128) \citep{dyrmann2016plant}} & Multiclass & \tabincell{l}{Combination\\ of 6 Public\\ Dataset}  & Y     &  \tabincell{l}{22\\ (crops + weeds)}  & -     & 86.2  & -     & -     & -     & - \\
    \midrule
    \tabincell{l}{Customized \\Residual Net\\(128X128) \citep{dyrmann2016plant}} & Multiclass & \multirow{2}*{\tabincell{l}{\\Grass-Broadleaf}} & Y       & \tabincell{l}{soil, soybean,\\ grass, broadleaf}  & -     & 89.1  & \multicolumn{1}{r}{66.3} & -     & -     & - \\
    AlexNet \citep{dos2017weed} & Multiclass & ~ & Y       & \tabincell{l}{soil, soybean,\\ grass, broadleaf}  & -     & 99.5  & -     & \multicolumn{1}{r}{99.5} & -     & - \\
    \midrule
    \tabincell{l}{K-means pre-training\\ and CNN \citep{tang2017weed}} & Multiclass & in-house & -       & \tabincell{l}{soybean,\\ 3 weeds}  & -     & 92.9  & -     & -     & -     & - \\
    \midrule
    VGGNet \citep{yu2019weed} & Multiclass & in-house & -     &  3 weeds & -     & \multicolumn{1}{l}{-} & \multicolumn{1}{r}{98.2} & \multicolumn{1}{r}{99.1} & \multicolumn{1}{r}{98.6} & - \\
    VGGNet \citep{yu2019weed} & Multiclass & in-house & -     &  3 weeds & -     & \multicolumn{1}{l}{-} & \multicolumn{1}{r}{98.6} & \multicolumn{1}{r}{93.4} & \multicolumn{1}{r}{95.6} & - \\
    \midrule
    VGGNet \citep{yu2019deep} & Multiclass & in-house & -     &  3 weeds & -     & \multicolumn{1}{l}{-} & \multicolumn{1}{r}{95.1} & \multicolumn{1}{r}{99.1} & \multicolumn{1}{r}{97.1} & - \\
    VGGNet \citep{yu2019deep} & Multiclass & in-house & -     &  3 weeds & -     & \multicolumn{1}{l}{-} & \multicolumn{1}{r}{93.7} & \multicolumn{1}{r}{99.9} & \multicolumn{1}{r}{96.7} & - \\
    \midrule
    PCANet \citep{xinshao2015weed} & Multiclass & in-house & -     &  91 weed seeds & -     & 91.0  & -     & -     & -     & - \\
    \midrule
    \kun{VGGNet \citep{zhuang2022evaluation}} & Multiclass & in-house & -     &  2 weeds & -     & \multicolumn{1}{l}{-} & \multicolumn{1}{r}{99.9} & \multicolumn{1}{r}{99.9} & \multicolumn{1}{r}{99.9} & - \\
    \bottomrule
    \end{tabular}
    }
    \begin{tablenotes}
        \footnotesize
        \item[*] Multi-label classification problem. 
    \end{tablenotes}
  \end{threeparttable}
  \label{tab:classification}%
\end{table*}

% Table generated by Excel2LaTeX from sheet 'Object Detection (2)'
\begin{table*}[htbp]
  \centering
  \caption{Summary of object detection based weed recognition studies. }
  \scalebox{0.85}{
    \begin{tabular}{lllrrrl}
    \toprule
    Architecture & Target Species & Modality & \multicolumn{1}{l}{Prec.} & \multicolumn{1}{l}{Recall} & $F_1$ Score & mIoU \\
    \midrule
    Faster R-CNN \citep{veeranampalayam2020comparison} & 5 weeds      & RGB   & 66.0  & 68.0  & 67.0  & \multicolumn{1}{r}{0.84} \\
    \midrule
    DetectNet (1224X1024) \citep{dyrmann2017roboweedsupport} & weeds      & RGB   & 86.6  & 46.3  & 60.3  & \multicolumn{1}{r}{0.64} \\
    \midrule
    YOLOv4 \citep{sharpe2020vegetation} & broadleaves, sedges, grasses     & RGB   & 100.0 & 91.0  & 95.0  & - \\
    \midrule
    DetectNet (1280X720) \citep{yu2019deep} & \textit{Poa annua}    & RGB   & 100.0 & 99.6  & 99.8  & - \\
    \midrule
    DetectNet (1280X720) \citep{yu2019deep} & \textit{Poa annua}    & RGB   & 100.0 & 99.8  & 100.0 & - \\
    \bottomrule
    \end{tabular}%
    }
  \label{tab:objectdetection}%
\end{table*}%

% Table generated by Excel2LaTeX from sheet 'Semantic Segmentation (2)'
\begin{table*}[htbp]
  \centering
  \caption{Summary of semantic segmentation based weed recognition studies.}
  \begin{threeparttable}
  \scalebox{0.75}{
    \begin{tabular}{lllllllrrll}
    \toprule
    Architecture & Dataset & \tabincell{l}{Target\\ Species} & \tabincell{l}{Public\\ Avail.}  & \multicolumn{1}{l}{\tabincell{l}{$F_1$\\ Score}} & Prec. & Recall & \multicolumn{1}{l}{mIoU} & \multicolumn{1}{l}{mAcc} & mAP   & AUC \\
    \midrule
    \multicolumn{11}{c}{\textbf{Multispectral (NIR \& RGB), Depth}}\\
    \midrule
    Customized U-Net \citep{brilhador2019classification} & Sugarbeets2016 & sugar beet, weeds, soil & Y   & 83.4  & -     & -     & \multicolumn{1}{l}{-} & \multicolumn{1}{l}{-} & -     & - \\
    DeepLabv3+ \citep{wang2020semantic} & Sugarbeets2016 & sugar beet, weeds, soil & Y     & \multicolumn{1}{l}{-} & -     & -     & 87.1  & \multicolumn{1}{l}{-} & -     & - \\
    \midrule
    \multicolumn{11}{c}{\textbf{Multispectral (NIR \& RGB)}}\\
    \midrule
    \tabincell{l}{Customized CNNs\\ (sNet + cNet) \citep{potena2016fast}} & in-house & crops, weeds, soil & -      & \multicolumn{1}{l}{-} & -     & -     & \multicolumn{1}{l}{-} & 92.0  & \multicolumn{1}{r}{97.4} & - \\
    \midrule
    Customized FCN \citep{lottes2018fully} & BONN  & crops, weeds, soil & Y   & 86.6  & \multicolumn{1}{r}{93.3} & \multicolumn{1}{r}{81.5} & \multicolumn{1}{l}{-} & \multicolumn{1}{l}{-} & -     & - \\
    \midrule
    Customized FCN \citep{lottes2018fully} & STUTTGART & crops, weeds, soil & -     & 92.4  & \multicolumn{1}{r}{91.6} & \multicolumn{1}{r}{93.5} & \multicolumn{1}{l}{-} &       & -     & - \\
    \midrule
    SegNet \citep{sa2017weednet} & Sugar beet/Weed & crops, weeds, soil & Y    & 80.0  & -     & -     & \multicolumn{1}{l}{-} & \multicolumn{1}{l}{-} & -     & \multicolumn{1}{r}{78.0} \\
    VGG-UNet~\citep{fawakherji2019uav}\tnote{*} & Sugar beet/Weed & crops, weeds, soil & Y   &       & -     & -     & \multicolumn{1}{l}{-} & 95.0  & -     & - \\
    \midrule
    \multicolumn{11}{c}{\textbf{RGB}}\\
    \midrule
    Customized CNN \citep{knoll2018improving} & in-house & carrots, weeds, soil & -     & 98.6  & \multicolumn{1}{r}{98.5} & -     & \multicolumn{1}{l}{-} & 97.9  & -     & - \\
    \midrule
    U-Net + VGG-16 \citep{fawakherji2019crop} & Sunflower & sunflower, weed, soil & -      & \multicolumn{1}{l}{-} & -     & -     & 80.0  & \multicolumn{1}{l}{-} & -     & - \\
    \midrule
    AgNet \citep{mccool2017mixtures} & CWFID & carrots, weeds, soil & Y       & \multicolumn{1}{l}{-} & -     & -     & \multicolumn{1}{l}{-} & 88.9  & -     & - \\
    MiniInception \citep{mccool2017mixtures} & CWFID & carrots, weeds, soil & Y      & \multicolumn{1}{l}{-} & -     & -     & \multicolumn{1}{l}{-} & 89.9  & -     & - \\
    \midrule
    Customized VGG-16 \citep{dyrmann2016pixel} & in-house & maize, weeds, soil & -   & \multicolumn{1}{l}{-} & -     & -     & 84.0  & 96.4  & -     & - \\
    \midrule
    \tabincell{l}{Customized CNN with\\ ResNet-10 Backbone \citep{li2019real}} & CWF-788 & cauliflower, weeds, soil & Y     & 98.0  & -     & -     & 95.9  & \multicolumn{1}{l}{-} & -     & - \\
    \midrule
    DeepLabv3+ \citep{wang2020semantic} & Oilseed Image & oilseed, weeds, soil & -     & \multicolumn{1}{l}{-} & -     & -     & 88.9  & \multicolumn{1}{l}{-} & -     & - \\
    \midrule
    VGG16-D \citep{mortensen2016semantic} & in-house & \tabincell{l}{equipment,soil,stump,\\ weeds,grass,radish,\\unknown} & -     & \multicolumn{1}{l}{-} & -     & -     & 55.0  & 79.0  & -     & - \\
    \midrule
    FCN-8s \citep{skovsen2019grassclover} & GrassClover & \tabincell{l}{grass, white clover, \\ red clover, weeds, soil} & Y      & \multicolumn{1}{l}{-} & -     & -     & \multicolumn{1}{l}{-} & \multicolumn{1}{l}{-} & -     & - \\
    \midrule
    SegNet \citep{lameski2017weed} & Carrot-Weed & carrots, weeds, soil & Y     & \multicolumn{1}{l}{-} & -     & -     & \multicolumn{1}{l}{-} & 64.1  & -     & - \\
    \midrule
    \tabincell{l}{SegNet \\(ResNet-50 Backbone)\\ \citep{asad2019weed}} & in-house & canola, weeds & -    & 99.3  & -     & -     & 82.9  & 99.5  & -     & - \\
    \bottomrule
    \end{tabular}%
    }
     \begin{tablenotes}
        \footnotesize
        \item[*] additional information involved from NVDI modality. 
      \end{tablenotes}
    \end{threeparttable}
  \label{tab:semanticsegmentation}%
\end{table*}%

\subsection{Weed Recognition Methods}
\label{sec:weedidmethod}

Existing studies on weed recognition can be organized into four categories in terms of the approaches they used: weed image classification, weed object detection, weed object segmentation or weed instance segmentation. \kun{Each approach represents a trade-off between algorithm complexity (i.e., speed of inference, training data difficulty) and the level of in-field recognition granularity that is provided as an output. We suggest that the selection of an approach will depend on the crop-weed combination, the weed control treatment scenario and the training and on-vehicle inference constraints. Table~\ref{tab:tb_approaches} provides an overview of these approaches from multiple perspectives such as computational cost and speed measured by floating-point operations per second (FLOPS), model size, power consumption, annotation intensity, recognition granularity and potential treatments. } Tables~\ref{tab:classification}, \ref{tab:objectdetection}, \ref{tab:semanticsegmentation} summarise the major studies of the first three categories, whilst instance segmentation based weed recognition has emerged recently. %has been focused very recently requiring more explorations in the future. 

\subsubsection{Weed Image Classification} 
This approach aims to achieve image-level weed recognition, determining of what species or crop/non-crop plants an image contains. An early deep learning based study \citep{dyrmann2016plant} devised a residual CNN for multi-class classification. On their proposed Grass-Broadleaf dataset which contains 10,413 RGB crop-weed images of 22 weed species, an accuracy 86.2\% was achieved. 
A variant PCA (Principal Component Analysis) network was proposed for classifying 91 classes of weed seeds using RGB images, and an accuracy 90.96\% was achieved. 
AlexNet was adopted to classify RGB images from the public dataset - Grass-Broadleaf \citep{dos2017weed}, which achieved an accuracy 99.5\%.
Although these results look promising, the plants or seeds were well segmented and the field or natural background information was limited, which could lead to failures under real field conditions.

More recently, a hybrid model of AlexNet and VGGNet was proposed. It was evaluated on a public plant seedling dataset containing RGB images of 3 crop species and 9 weed species at an early growth stage and achieved an accuracy 93.6\%. 
Classifying three weed species including \textit{Euphorbia maculata}, \textit{Glechoma hederacea} and \textit{Taraxacum officinale} growing in perennial ryegrass 
was studied \citep{yu2019weed} using VGGNet on RGB images. It achieved F-scores 98.6\% and 95.6\% for two independent test sets collected from two fields with different locations. 
A similar study was also conducted to classify three other species of weeds growing in perennial ryegrass including \textit{Hydrocotyle spp}, \textit{Hedyotis cormybosa} and \textit{Richardia scabra}~\citep{yu2019deep}. 
%A similar study classifies three other species of weeds growing in perennial ryegrass including \textit{Hydrocotyle spp}, \textit{Hedyotis cormybosa} and \textit{Richardia scabra} using VGGNet with RGB images \citep{yu2019deep}. It achieved 97.1 and 96.7 of the F-Scores for two independent test sets, which were collected from different locations. 
Another study identified cephalanoplos, digitaria, bindweed and soybean in RGB images by introducing a CNN based on LeNet-5~\citep{ciresan2011convolutional} with a K-means clustering for unsupervised pre-training, which achieved an accuracy 92.9\% \citep{tang2017weed}. 
To further advance weed image classification in complex environments, a public dataset, namely DeepWeeds \citep{olsen2019deepweeds}, was constructed by acquiring RGB images in remote and extensive rangelands with rough and uneven terrains. The baseline accuracy 95.7\% was achieved by a ResNet-50 for multi-label classification. A simplified DenseNet, namely DenseNet-128-32, was explored to reduce the computational cost and inference time, while keeping the performance comparable to that of the original DenseNet model \citep{lammie2019low}. \kun{More recent field-based studies have found image classification (including AlexNet, DenseNet, ResNet and VGGNet) outperformed object detection networks for the recognition of broadleaved seedlings in wheat \citep{zhuang2022evaluation}. All image classification networks tested had F1 scores above 0.99. }
%It achieved an accuracy of 90.1\% on the DeepWeeds dataset and significantly alleviated the inference costs. 

Recently, a few fine-grained architectures were explored to improve weed image classification performance. 
By introducing graph-based image representation, a graph weeds net achieved the state-of-the-art accuracy 98.1\% on the DeepWeeds dataset, which formulated global and fine-grained weed characteristics with GCNs \citep{hu2020graph}.
Another study also investigated the graph mechanisms \citep{jiang2020cnn}, in which GCN-ResNet-101 was proposed and the accuracy values varied from 96.5\% to 98.9\% on 4 public RGB datasets. %The accuracy values were varied from 96.5\% to 98.9\%. 

Deep unsupervised learning was explored in a recent weed study \citep{dos2019unsupervised}, which explored two methods: joint unsupervised learning of deep representations and image clusters (JULE) and deep clustering for unsupervised learning of visual features (DeepCluster). They adopted the CNN outputs as features for a clustering algorithm and specified pseudo labels for samples based on the clustering results. 
%Annotations were not involved in this study and two benchmark datasets with RGB images were evaluated. 
As reported, the DeepCluster method achieved an accuracy 70.6\% with a VGG-16 backbone on the DeepWeeds dataset, and an accuracy 99.5\% with an AlexNet backbone on Grass-Broadleaf. %Note that the latter dataset is generally easier for classification due to the well segmented plants and fewer classes. 

Besides RGB images, multispectral imaging has also been investigated. Different sizes of multispectral images with a CNN involving 4 convolution layers were explored \citep{farooq2018analysis}. By varying the input size from $125\times 125$ to $500\times 500$, the accuracy varied from 86.3\% to 94.7\% on the UNSW Hyperspectral Weed Dataset.  FCNN-SPLBP combined CNN and superpixel based local binary pattern feature extraction \citep{farooq2019multi}, which was evaluated on two public datasets: an accuracy 89.7\% for $100\times 100$ pixel images on the UNSW Hyperspectral Weed Dataset and an accuracy 96.4\% on the sugar beet/weed dataset. \kun{With the coarse, whole-image granularity of image classification, the method is more likely to be useful in coarse weed control scenarios, such as spot spraying \citep{calvert2021robotic}. }

\subsubsection{Weed Object Detection} 
%this approach detect weeds and generates bounding boxes of the detected weed plants. 
\kun{Moving beyond the whole-image level of understanding, object detection provides bounding box coordinates of detected weeds. The additional contextual information allows targeting whole weed plants amongst crops, individual leaves, or other plant components. For instance, detecting individual leaves was found to be more effective than whole plant detection in strawberry raised beds  \citep{sharpe2019detection}.} Existing weed object detection methods are mainly based on generic object detection methods. 
%Object detection models such as Faster R-CNN, DectectNet and YOLO were introduced for weed identification tasks using RGB images. 
In \citep{dyrmann2017roboweedsupport}, DetectNet was used with an mIoU 64.0\% and an F-score 60.3\% on an in-house dataset containing 1,427 RGB images. 
Another study \citep{yu2019deep} with DetectNet achieved F-scores 99.8\% and 100.0\% for two different environments in detecting a weed species, namely \textit{Poa annua}. 
In \citep{veeranampalayam2020comparison}, Faster R-CNN achieved an mIoU 84.0\% and an F-score 67.0\% in detecting waterhemp (\textit{Amaranthus tuberculatus}), Palmer amaranth (\textit{Amaranthus palmeri}), common lambsquarters (\textit{Chenopodiam album}), velvetleaf (\textit{Abutilon theophrasti}), and foxtail species such as yellow and green foxtails on an in-house dataset containing 450 augmented RGB images. 
YOLOv3 was used to detect broadleaves, sedges and grasses \citep{sharpe2020vegetation} and achieved an F-score 95.0\%. 

\subsubsection{Weed Semantic Segmentation} 
%this approach can be viewed as a pixel-wise classification problem, which predicts the category of each pixel. 
\kun{With pixel-level granularity, semantic segmentation of weeds provides greater detail than object detection. The approach is more suited to precision weed control methods such as laser weeding, which must only hit the plant if they are to be effective.} An intuitive approach for weed semantic segmentation is based on a two-stage scheme. Two CNNs were devised in~\citep{potena2016fast}, namely sNet and cNet, in which the first stage generated segmented objects and the second stage predicted the class of each object \citep{potena2016fast}.
The method were applied to multispectral images to identify the regions of crops, weeds and soil with a mean accuracy 92.0\% and an mAP 97.4\%. 
Another study adopted a conventional algorithm using HSV-colour room vegetation index method for segmentation and a CNN for classification \citep{knoll2018improving}. It achieved an F-score 98.6 and a mean accuracy 97.9\% on an in-house RGB image dataset containing carrots, weeds and soil. 
%chen2018encoder DeepLabv3+

FCNs were investigated in pursuit of end-to-end solutions which treat the segmentation and classification within one neural network. 
In \citep{dyrmann2016pixel,mortensen2016semantic}, the last FC layer of a VGG-16 was replaced as a deconvolutional layer. The modified VGG-16 was evaluated on two RGB datasets: one to segment maize, weeds and soil and an mIoU 84.0\% and a mean accuracy 95.4\% was achieved; on another to segment equipment, soil, stump, weeds, grass, radish and unknown categories, a mean accuracy 79.0\% was achieved.
%A similar strategy was used in \citep{dyrmann2016pixel}, a customized VGG-16 network modified the last fully connected layer of the original VGG-16 as a deconvolutional layer with shortcuts. It was evaluated to segment maize, weeds and soil from RGB images and an mIoU 84.0\% and a mean accuracy 95.4\% were achieved. 
%In \citep{mortensen2016semantic}, by converting the last fully connected layer of a VGG-16 to deconvolutional layer, a VGG-16D was obtained to segment equipment, soil, stump, weeds, grass, radish and unknown categories in an RGB image, by which a mean accuracy 79.0\% was obtained. 
A FCN with a DenseNet backbone \citep{lottes2018fully} was evaluated to identify crops, weeds and soil in multispectral images and achieved F-Scores 86.6\% and 92.4\% on two datasets collected from sugar beet fields in two different locations. 
Two FCN-8s were trained to segment RGB images: the first one recognized grass, clover, weeds and soil, and the latter one recognized fine-grained clover species including white clover and red clover \citep{skovsen2019grassclover}. It achieved an mIoU 0.55 on its proposed GrassClover dataset. 

In addition to using simple FCNs, recent studies tended to explore FCNs with additional mechanisms that are beneficial for segmentation tasks. In \citep{lameski2017weed}, a SegNet with VGG-16 backbone achieved a mean accuracy 64.1\% on a carrot-weed dataset containing RGB images of carrots, weeds and soil. In~\citep{sa2017weednet}, a public multispectral dataset - Sugar beet/Weed - was proposed to identify crops, weeds and soil, and a SegNet with a VGG-16 backbone achieved an F-Score 80.0\% and an AUC 78.0\% by evaluating the crops and the weeds predictions within a binary pixel-wise classification scheme. 
In~\citep{asad2019weed}, a SegNet with a ResNet-50 backbone was adopted to identify canola and weeds in RGB images using a pre-processing step to remove backgrounds, which achieved an mIoU 82.9\% and a mean accuracy 99.5\%. 
A Bonnet framework \citep{milioto2019bonnet} to segment sunflower, weed and soil in RGB images achieved an mIoU 80.0\% \citep{fawakherji2019crop}. A customized U-Net with different data augmentation strategies was investigated for the CWFID dataset \citep{brilhador2019classification}, which achieved an F-score 83.4\%. A VGG-UNet was evaluated for the Sugar beet/Weed dataset and achieved a mean accuracy 95.0\% \citep{fawakherji2019uav}. 
DeepLab-v3 was evaluated on Sugarbeet2016 containing multispectral and depth data and an in-house RGB oilseed dataset, which achieved mIoU values of 87.1\% and 88.9\%, respectively~\citep{wang2020semantic}.

Lightweight models aiming for efficient segmentation were explored. In \citep{mccool2017mixtures}, light models were mixed together with the guidance of a large and accurate model Inception-V3. On the CWFID dataset,
compared to the Inception-V3 with an accuracy 93.9\% and 0.12 fps during the inference, 4 mixed lightweight models %(CNN models similar to AlexNet, of which the parameters were 25 times less than the Inception-V3) 
achieved an accuracy 90.3\% with 1.83 fps. % during the inference. 
%AgNet and MiniInception network were evaluated on a public dataset containing RGB images to segment carrots, weeds and soil \citep{mccool2017mixtures} with mean accuracy values 88.9\% and 89.9\% for AgNet and MiniInception network, respectively. 
%Note that the results were slightly lower compared to the conventional model \hl{[]which conventional model? why?]}. For example, adapted InceptionV3 was able to achieve a mean accuracy 93.9\%. 
A customized CNN using a ResNet-10 backbone \citep{li2019real} with side outputs and short connections for multi-scale feature fusion achieved an F-score 98.0\% and an mIoU 0.959 on the CWF-788. 

\subsubsection{Weed Instance Segmentation} 
\kun{Weed instance segmentation provides the highest-level granularity for weed recognition, with information on both pixel class and to which each weed the pixel belongs. As with semantic segmentation, the most likely use-case for the approach is with highly targeted weed control treatments. Understanding which weed to target rather than targeting every weed pixel, would greatly improve efficiency.} There have been a few attempts at deploying instance segmentation algorithms for weed recognition. A recent study adopted Mask R-CNN for field RGB images of two crop species and four weed species \citep{champ2020instance}. Further explorations for this approach are needed by considering the perspectives such as the improvements on different growth stage and small size plants. 

\section{Discussion}
\label{sec:discussion}
In this section, based on weed control challenges and the recent developments of deep learning techniques, 
%the potential perspectives of the improvements for weed identification are discussed for future studies in this realm. In detail, the discussion mainly covers 
we discuss the challenges and opportunities for further advancing weed recognition research from the following aspects: fine-grained learning, real-time inference, explainability, weakly-supervised / unsupervised learning, and incremental learning. 

\subsection{Fine-grained Learning}

As reviewed in Section \textit{Weed Recognition Methods}, most of the existing weed recognition methods were based on general deep architectures ignoring the challenge caused by the strong similarities of crops and weed species.  
%developed from deep models that were proposed for generic objects (e.g., car) of which inter-class variations are generally more distinctive than intra-class variations.
%However, weed identification imposes unique challenges due to the strong similarity between crops and weeds or between different weed species.
%This requires advanced deep learning techniques which are able to undertake the learning task at a finer granularity. 
Recently, 3 major categories of fine-grained deep methods were explored to address this challenge. 

%Current weed management studies usually use architectures devised in a general manner, which does not take the domain knowledge of weed identification into consideration. However, in many problems, the intra-class variation can be higher than the inter-class variation for the images of crops and weeds. For example, target weeds and desirable plants look largely similar in weed classification tasks, which increases the difficulties to obtain a promising deep learning model.
%To address this issue, deep architectures involving the domain knowledge were explored for various tasks and we first review some of them to present their basic ideas. 

%\subsubsection{Patch-based Methods}

\textbf{Patch-based methods} based on the fact that the fine-grained details often occur at a local-level. With the patterns collected from each region, fused or aggregated methods can be used to compute the final outputs. 
For example, regional CNN based features can be collected according to the key points of human poses for fine-grained action recognition~\citep{hu2019vision}. %For fine-grained animal species classification, body parts and their relations were focused~\citep{peng2018object}. %Similarly, another study tries to construct recurrent attention over a series of patches to look closer to the key characteristics for fine-grained classification~\citep{fu2017look}. 

\textbf{High-order pooling based methods} were introduced to address fine-grained tasks as well, which did not require explicit patch proposals (e.g. \citep{zheng2019looking}). In particular, for a given convolutional feature map $\mathbf{X}\in \mathbf{R}^{c\times wh}$, the bilinear pooling can be computed by $\mathbf{X}^\top \mathbf{X}$; the trilinear pooling can be computed by $(\mathbf{X}^\top \mathbf{X})\mathbf{X}$. The pooled vector can be used as the input of the subsequent layer of the network. 
The relations between the high-order pooling methods and the patch-based methods can be explained from the perspective of the attention \citep{kim2018bilinear}. Both of these methods result in focusing on the critical regions to collect efficient deep representations for their associated tasks.

\textbf{Regularization based methods}
are based on that the intra-class difference could be higher than the inter-class difference for fine-grained modelling, to introduce regularization terms for loss and drive the optimization to focus on learning fine-grained patterns. 
For example, in \citep{DubeyGGRFN17}, %training-with-confusion, which includes
pair-wise confusion and entropic confusion was introduced to construct its loss function. 
%Regularization of the loss functions, which push the optimization focusing on the subtle difference instead of the intra-class difference, was proposed as well based on the fact that the inter-category difference is usually less than the intra-class difference \citep{DubeyGGRFN17}. 

%It is interesting to know how these fine-grained architectures performed for weed management. However, limited studies involved such techniques. Very recently, graph weed net investigated patch-based fine-grained classification on DeepWeeds dataset, which improved the accuracy 95.3\% of DenseNet to 98.1\% \citep{hu2020graph}.

Such fine-grained deep models provide a great opportunity to advance weed recognition by taking the domain knowledge into account. For example, a weed can be decomposed into meaningful regions, such as leaves and stems. In our recent work, a patch-based GNN \citep{hu2020graph} was proposed towards fine-grained weed classification,  which achieved an accuracy 98.1\% on the DeepWeeds dataset, compared with the accuracy 95.3\% of DenseNet.

\subsection{Real-time Inference}

While most of the weed recognition studies demonstrated promising performance using deep learning techniques, these deep networks often contain a huge number of parameters. It leads to three major issues in regard to efficiency, memory consumption, and power consumption for deployment. 
%In this subsection, recent model compression and acceleration techniques to alleviated these issues are discussed. We also discuss the existing studies aiming for efficient weed identification by exploring these techniques as listed in Table~\ref{tab:realtime}. 
Intuitively, as indicated in~\citep{cheng2017survey}, lightweight models (e.g. MobileNet \citep{howard2019searching}) can be devised by using mechanisms such as parameter pruning, low-rank factorization, transferred/compact convolutional filters. 
%For example, the commonly used depth-wise separable convolution, is a form of factorized convolution, which factorizes a general convolution into a depth-wise convolution (a single filter to each input channel) and a point-wise convolution (1X1 convolution) to reduce parameters in a model.
%Intuitively, investigating mechanisms to involve less model parameters helps to reduce computational costs. 
%Following these mechanisms, a number of deep architectures for mobile and embedded vision applications were proposed such as MobileNet \citep{howard2019searching}%, %SqueezeNet \citep{iandola2016squeezenet}, 
%ShuffleNet \citep{Zhang_2018_CVPR}
%and 
%Xception \citep{Chollet_2017_CVPR}, 
%EfficientNet \citep{tan2019efficientnet}. 
In particular, using a Google Pixel 3 device with one-thread on a single large core, the inference time of MobileNet (V3) for image classification on ImageNet achieved a top-1 accuracy 65.4 and inference latency 11.7 ms. Note that these lightweight models can also be treated as backbones for object detection and segmentation. For example, SSDLite with MobileNet (V3) Small backbone achieved an inference latency of 43 ms and an mAP 16.1 on COCO test set; MobileNet (V3) based segmentation achieved an mIoU 69.4 with an inference time of 1.03s for an input image with resolution of $1024\times 2048$. 
For weed recognition, a ResNet-10 was proposed as backbone and introducing side outputs and short connections for multi-scale feature fusion. It achieved an mIoU 95.9 and an F-score 98.0, while the average inference latency is around 180ms on an Nvidia Jeston TX2 \citep{li2019real}. 

\begin{table*}[htbp]
  \centering
  \caption{Examples of real-time inference latency (ms) and performance for weed detection. }
  \scalebox{0.85}{
    \begin{tabular}{lllll}
    \toprule
    Architecture & Modality & Latency & Performance Metrics & Device \\
    \midrule
    \tabincell{l}{Customized CNN with ResNet-10 Backbone \citep{li2019real}} & RGB   & \multicolumn{1}{r}{180} & 95.9 mIoU & Jetson TX2 \\
    Binarized DenseNet-128-32 \citep{lammie2019low} & RGB   & \multicolumn{1}{r}{1.539} & 90.1 Acc & Terasic DE1-SoC \\
    Mixture of lightweight models \citep{mccool2017mixtures} & Multispectral & 546-934 & 90.0 Acc & GeForece Titan X \\
    \bottomrule
    \end{tabular}%
    }
  \label{tab:realtime}%
\end{table*}

In addition to devising lightweight architectures, methods such as quantization and knowledge distillation are devised for any existing models with less parameters while providing comparable performance as the complex models (e.g. ResNet-50 \textit{vs.} ResNet-152).  
Quantization methods reduce the number of bits to represent the parameters in a model. In particular, binarization only saves one bit for parameter, which significantly reduces the memory consumption and computational cost \citep{qin2020binary}. A binarized DenseNet-128-32 was implemented by FPGA (Terasic DE1-SoC) for weed detection gaining an accuracy 88.91\%~\citep{lammie2019low}. It was slightly lower than a general DenseNet but obtained a very fast average inference latency 1.539ms. 
Knowledge distillation follows a similar way in which human beings learn, which contains one or more large pre-trained teacher models and a small student model. It aims to obtain an efficient student model which mimics and performs comparably to the the teacher models. A distillation loss penalizes the difference between the outputs from the teacher and the student models. 
A weed recognition study followed this scheme to obtain a few lightweight models for semantic segmentation~\citep{mccool2017mixtures}. Mixing these lightweight models achieved an accuracy 90.0\% and the inference latency between 934ms to 546ms by using an Nvidia GeForce Titan X graphics card.

\subsection{Weakly-supervised \& Unsupervised Learning}

As manually collected supervision information for weed dataset can be resource expensive, %with increasingly difficult from image-level annotation to object-level annotation and pixel-level annotation. 
%Whereas it is relatively easier to collect image-level or bounding-box level ground truth information, such annotation can only be done through trained experts with domain expertise, %unlike annotating generic objects, 
%which makes annotating weed images more challenging. 
%To alleviate such challenges, 
weakly-supervised and unsupervised learning algorithms are needed for weed recognition. 
For weakly-supervised learning, it is expected that weed object detection or even weed segmentation can be conducted when only using image-level annotation. %, which is highly attractive and beneficial. 
%To obtain an accurate model using deep learning, it often requires a huge amount of samples during the training stage due to the model complexity. For weed identification, the acquisition of images can be automatically finished by involving acceptable human efforts. However, the annotations of these images can be costly, which requires well-trained experts to identify the labels for different levels (i.e., image, object, or pixel) as the supervision information for training. 
%Therefore, it is attractive to devise methods, which can be benefit from a large amount of unlabelled samples. According to the extent of the supervision involved, these methods can be categorized as semi-supervised or unsupervised learning methods. 
%For semi-supervised learning, it is assumed that the unlabelled data contains the underlying distribution information of the population, which could be useful, for example, to construct the decision boundary of a classification problem.
For unsupervised learning, deep clustering and domain adaptation can be conducted. Deep clustering categorizes similar samples into one cluster in line with some similarity measures on their deep representations~\citep{min2018survey}. 
%It often involves a clustering loss, which depends on a data-driven manner rather than supervised labelling. 
An application of deep clustering is that pre-training a neural network with a large unlabelled dataset and further fine-tuning on a small labelled dataset.  
Domain adaptation solves the problem that the training samples and testing samples following different distributions. This could be the case, for example, two datasets for the same species are from different locations. Therefore, unsupervised domain adaptation 
handles situations where a network is trained on labeled data from a source domain and unlabeled data from a related but different target domain. Readers can refer to~\citep{wilson2020survey} for more details. 

Note that existing deep learning based weed recognition methods have not adequately explored this realm to use the unlabelled samples. Until very recently, deep clustering was investigated for weed image classification \citep{dos2019unsupervised}, in which a VGG-16 based DeepCluster network achieved an accuracy 70.6\% on the DeepWeeds dataset. In \citep{hu2020graph}, a GraphWeedsNet involved a weakly-supervised learning strategy, namely multi-instance learning, and used image-level annotations to provide approximate locations of weed plants. 

\subsection{Explainable Learning}

Deep learning shows a black-box nature, since it is difficult to understand and interpret the relations between the inputs and the outputs. % of deep models, compared to other conventional machine learning methods. 
However, explainability is of great importance for building trust between models and users to eventually facilitate the model adoption. % and monitoring. %of deep learning techniques and for monitoring the deployment of deep models. %can be important, which can instruct the use of the trained models. 
As summarized in~\citep{xie2020explainable}, there are three major approaches in pursuit of the explainability of deep learning: 1) visualization methods identify the most important parts of an input, which highly influence the results; 2) model distillation involves conventional machine learning models, most of which have clear statistical explanations and indications, to mimic the behavior of trained deep models; 
3) intrinsic methods integrate mechanisms (e.g., the attention mechanisms).%to deep learning models and their training
%, by which the model performance is often improved as a bonus.

Explainable learning has been seldom investigated for deep learning based weed recognition, although it has the potentials to provide further insights. Recently, graph weeds net was proposed with its graph mechanism, which treats the regions of an input image as graph vertices, to analyze the critical regions \citep{hu2020graph}. %The graph mechanism can be viewed as an attention module indicate which regions dominantly contribute to the predictions. 
As it usually takes more efforts for object detection or segmentation than image classification, such explainable learning approach also provides an opportunity to take less effort to focus on the critical objects within an image. Furthermore, the critical regions are obtained without regional annotations, which can be viewed as a weakly-supervised learning requiring less human efforts. 

\subsection{Incremental Learning}

Most of the existing weed recognition methods assume that a trained network will only deal with fixed target species, which are available during the training.
As a result, when new species of interests are emerged, it is generally expected that the deep model needs to be re-trained with a new training set. To address this time-consuming and inflexible scheme, incremental learning is proposed to extends a trained model for new classes without the re-training from scratch.
Note that the training samples of existing species are often not stored with a high volume due to storage limitations, whilst samples of incremental species could be adequate. Hence, incremental learning mainly addresses this imbalance nature when obtaining a new model based on an existing model. 

To conduct incremental learning, 4 major approaches were investigated \citep{de2019continual}. 1) retaining a subset of the old training data in line with a budget. %For example, the exemplars per class, which best approximate the mean of each class in the feature space can be selected. 
2) The distributions of the old dataset can be stored as the parameters of a generative model, which can produce unlimited samples during the incremental training. 3) Parameter isolation-based methods aim to prevent any possible forgetting of the previous tasks when no constraints on the model size. In general, for different species, it can use different model parameters to conduct the classification. 4) Regularization techniques prevent forgetting previous knowledge.

Recently, AgroAVNET explored the chance for incremental learning on the plant seedling dataset~\citep{chavan2018agroavnet} and the accuracy achieved 91.35 for 12 species, compared to 93.64 from a general re-training. 
It followed a very straightforward way without fully exploiting the incremental learning, which froze the convolution layers trained by the original dataset and re-trained the FC layers only. %Currently, the incremental learning has not been fully exploited yet. 

\subsection{Large Scale Datasets}
Large scale datasets are essential for developing high performance and robust deep learning models. 
For example, ImageNet~\citep{krizhevsky2012imagenet}, which contains 15 million labeled images belonging to roughly 22,000
categories, has played a significant role in advancing deep learning based vision tasks. 
However, as summarized in Section \textit{Weed Data}, most of the existing weed datasets contain images of a small number of classes. In addition, those images were collected under limited scenarios, such as one growth stage and one light condition. 
This has limited the development of advanced methods applicable to a large variety of fields and prevents the translation towards commercial adoption. 
Therefore, constructing large scale datasets with diverse and complex conditions in the context of practical deployment can be highly demanded.% to further advance the field of weed identification towards enabling commercial use of novel weed control techniques. 

\section{Conclusion}
\label{sec:conclusion}

In this paper, we reviewed the recent progresses in the field of deep learning based weed recognition and discussed the challenges and opportunities for future research. 
After introducing the fundamentals of deep learning techniques, we provided a systematical review from three aspects: research data, evaluation, and weed recognition methods. % which are categorized into weed image classification, weed object detection, weed semantic segmentation and weed instance segmentation in terms of the identification granularity from image level to bounding box level and pixel level. 
There have been more than 10 public datasets collected through different modalities and many weed recognition methods have been reported across different research disciplines due to the inter-disciplinary nature of this topic. 
It is also noticed that most existing weed recognition methods were proposed by using the architectures developed for generic deep learning problems. \kun{Given the substantial differences in output granularity, the selection of specific recognition approaches should be governed by the in-field weed control treatment scenario. Where highly precise control methods are needed or where occlusion may reduce the effectiveness of coarser approaches, a trade-off may required in the complexity of architecture selected and hence complexity of training and deploying such an architecture in the field.}
Finally, we discussed the challenges and opportunities in terms of 5 different learning techniques and large scale dataset. 
Overall, deep learning based weed recognition has gained increasing interest from different research communities and we feel that large scale datasets are strongly needed to bring this research direction to a new level. 
%In this paper, deep learning was reviewed for weed identification tasks, which have gained great success for many visual-based applications. The deep learning techniques, especially related to the weed detection, were introduced including a number of the state-of-the-art architectures and mechanisms. 
%More than 30 papers focusing on deep weed identification were investigated, which were based on three approaches: image classification, object detection and semantic segmentation. In addition, as a deep learning model is often trained with a large volume dataset, to facilitate the research in this realm, a number of public weed related datasets were discussed in this review. 
%Besides summarizing these existing studies, the aim of this review also motivates researchers to explore the potential from several perspectives, which facilitate the learning abilities of the models and the applications for field use. 

\section*{Acknowledgement}
Funding: This work was supported by the GRDC (Grains Research and Development Corporation) [grant number 9177493].

%\begin{acknowledgements}
%\end{acknowledgements}

\section*{Conflict of interest}
The authors declare that they have no known competing financial interests or personal relationships that could influence the work reported in this paper.

\bibliographystyle{spbasic}      % basic style, author-year citations
\bibliography{ref}   % name your BibTeX data base

\end{document}